\documentclass[10pt,journal,compsoc]{IEEEtran}
\usepackage{amsmath,amsfonts}
\usepackage{algorithmic}
\usepackage{algorithm}
\usepackage{array}
\usepackage[caption=false,font=normalsize,labelfont=sf,textfont=sf]{subfig}
\usepackage{textcomp}
\usepackage{stfloats}
\usepackage{url}
\usepackage{verbatim}
\usepackage{graphicx}
\usepackage{cite}

\usepackage{_mathalias}
\usepackage[dvipsnames]{xcolor}
\usepackage{amsmath}
\usepackage{bbm}
\usepackage{color,colortbl}
\usepackage{enumitem}
\usepackage{booktabs}
\usepackage{nicefrac} 
\usepackage{amsfonts} 
\usepackage{multirow}
\usepackage{pifont}
\DeclareMathOperator*{\topk}{\operatorname{top-k}}  % declare top-k operator

\begin{document}

% commands
\newcommand{\inlineColorbox}[2]{\begingroup\setlength{\fboxsep}{1pt}\colorbox{#1}{\hspace*{2pt}\vphantom{Ay}#2\hspace*{2pt}}\endgroup}

% general shorts
\newcommand{\aka}{\textit{a}.\textit{k}.\textit{a}.}
\newcommand{\eg}{\textit{e}.\textit{g}.}
\newcommand{\ie}{\textit{i}.\textit{e}.}

% special shorts
\newcommand{\taskName}{\textcolor{black}{Vocabulary-free Image Classification}}
\newcommand{\taskShort}{\textcolor{black}{VIC}}
\newcommand{\taskSName}{\textcolor{black}{Vocabulary-free Semantic Segmentation}}
\newcommand{\taskSShort}{\textcolor{black}{VSS}}
\newcommand{\taskExpandName}{\textcolor{black}{Vocabulary-free Image Classification and Semantic Segmentation}}
\newcommand{\taskExpandShort}{\textcolor{black}{VICSS}}
\newcommand{\methodName}{\textcolor{black}{Category Search from External Databases}}
\newcommand{\methodShort}{\textcolor{black}{CaSED}}

\newcommand{\methodEShort}{\textcolor{black}{UpperCaSED}}

\newcommand{\methodSName}{\textcolor{black}{Dense Category Search from External Databases}}
\newcommand{\methodSShort}{\textcolor{black}{DenseCaSED}}

% colors
\definecolor{BlueEnrico}{RGB}{60, 151, 151}
\definecolor{RedEnrico}{RGB}{252, 100, 100}
\definecolor{GreenEnrico}{RGB}{80, 201, 80}
\definecolor{OrangeEnrico}{RGB}{252, 169, 100}
  % import macros

\title{Vocabulary-free Image Classification and Semantic Segmentation}

\author{Alessandro Conti, Enrico Fini, Massimiliano Mancini, Paolo Rota, Yiming Wang, Elisa Ricci
\thanks{A. Conti, M. Mancini, P. Rota, and E. Ricci are with the University of Trento (Trento, Italy), Y. Wang and E. Ricci are with Fondazione Bruno Kessler (Trento, Italy). E. Fini was affiliated with the University of Trento and is now at Apple. The work was done before joining Apple. Corresponding author: A. Conti (alessandro.conti-1@unitn.it).}
}

\maketitle

\begin{abstract}
Large vision-language models revolutionized image classification and semantic segmentation paradigms. However, they typically assume a pre-defined set of categories, or vocabulary, at test time for composing textual prompts. This assumption is impractical in scenarios with unknown or evolving semantic context. Here, we address this issue and introduce the Vocabulary-free Image Classification (VIC) task, which aims to assign a class from an unconstrained language-induced semantic space to an input image without needing a known vocabulary. VIC is challenging due to the vastness of the semantic space, which contains millions of concepts, including fine-grained categories. To address VIC, we propose Category Search from External Databases (CaSED), a training-free method that leverages a pre-trained vision-language model and an external database. CaSED first extracts the set of candidate categories from the most semantically similar captions in the database and then assigns the image to the best-matching candidate category according to the same vision-language model. Furthermore, we demonstrate that CaSED can be applied locally to generate a coarse segmentation mask that classifies image regions, introducing the task of Vocabulary-free Semantic Segmentation. CaSED and its variants outperform other more complex vision-language models, on classification and semantic segmentation benchmarks, while using much fewer parameters. Code is available at \url{https://github.com/altndrr/vicss}.
\end{abstract}


\section{Introduction}
\label{sec:intro}

Large-scale Vision-Language models (VLMs)~\cite{clip,yu2022coca,li2022blip} have revolutionized the field of computer vision, connecting multimodal information in an unprecedented manner. One peculiar aspect of these models is their zero-shot transfer capabilities: for instance, CLIP~\cite{clip} showed outstanding classification results in multiple datasets, even if not being explicitly trained {for the task at hand}. This lead to extending VLMs to other discriminative tasks, such as semantic segmentation~\cite{liang2023open,xu2023side,yu2023convolutions} or object detection~\cite{ma2023codet,kuo2022f}, where their multimodal nature allows to perform such tasks in an ``open-vocabulary'' manner, \ie, where the (finite) set of categories can be dynamically defined by the user.

\begin{figure*}[t!]
\centering
\begin{tabular}{ccc}
 \includegraphics[height=0.25\linewidth]{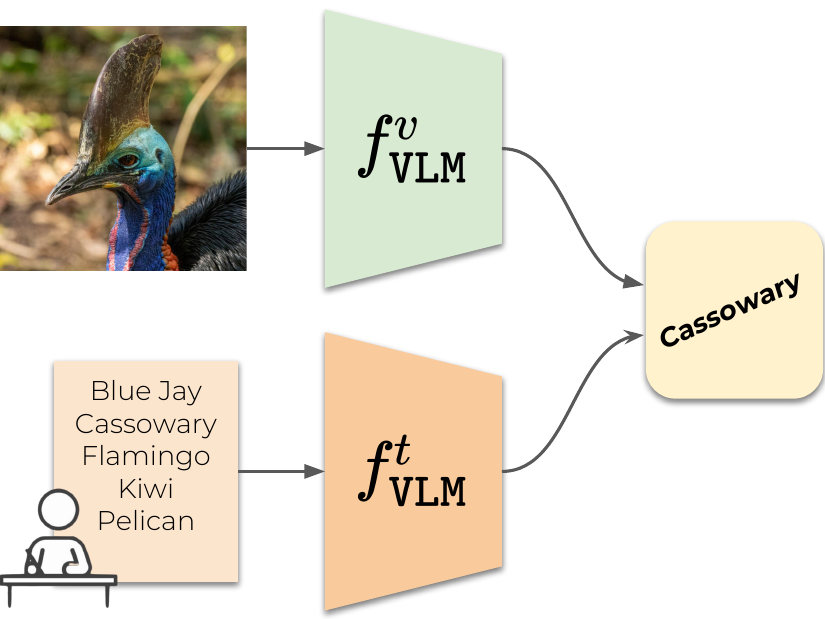} & & \includegraphics[height=0.25\linewidth]{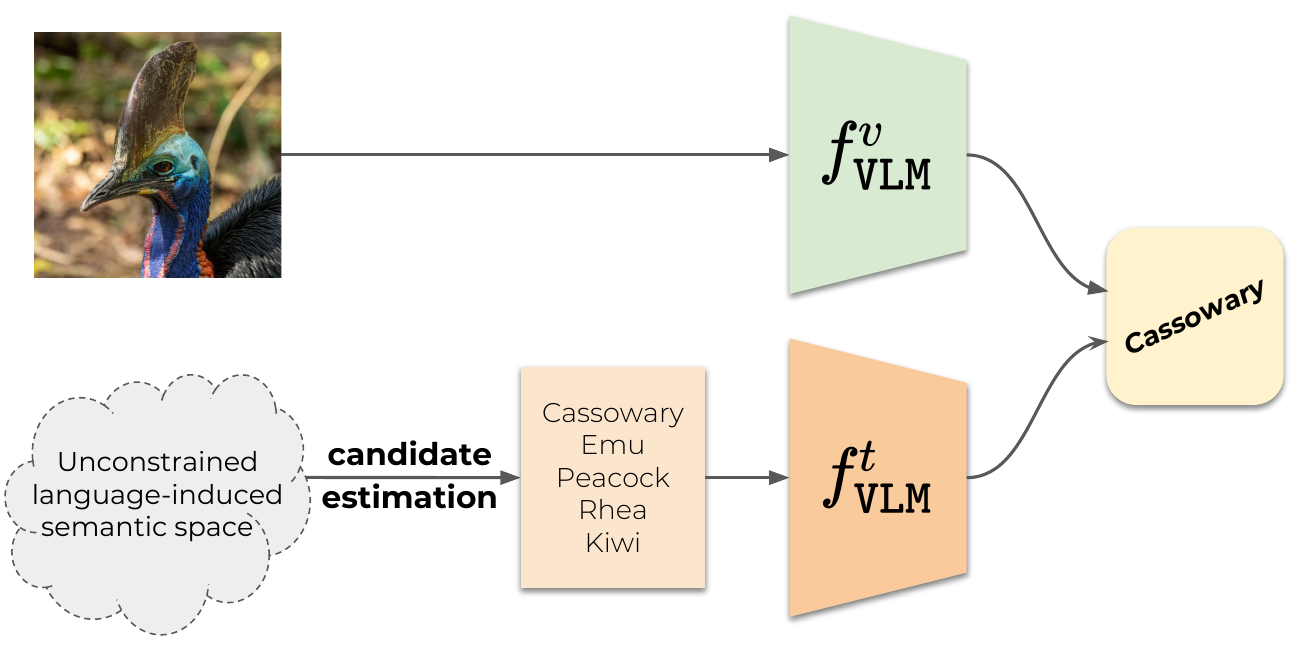} \\
 (a) VLM-based classification & & (b) \taskName \\
\end{tabular}
\caption{Vision-Language Model (VLM)-based classification (a) assumes a pre-defined set of target categories, \ie, the vocabulary, while our novel task (b) lifts this assumption by directly operating on the unconstrained language-induced semantic space.
$f^v_\mathtt{VLM}$ and $f^t_\mathtt{VLM}$ denote the pre-trained vision and text models of a VLM, respectively. In this work, we also extend this paradigm for the task of semantic segmentation.}
 \label{fig:vic_teaser}
\end{figure*}

In this paper, we aim to challenge the latter assumption and perform classification tasks with VLMs \textit{without} a set of target categories (\ie, the vocabulary) pre-defined by {the user} (see Fig.~\ref{fig:vic_teaser}).  
This has many practical advantages as this lack of priors often arises when working with autonomous agents in unconstrained environments. At the same time, it inherits various challenges as (i)  the search space encompasses all possible existing semantic concepts, even very fine-grained ones that are difficult to discriminate {and possibly ambiguous}; (ii) we need classification models that do not rely on vocabulary-aware supervision, to avoid potential biases on the sub-part of the vocabulary within the training data. We name this task \taskName\ (\taskShort).

Our approach exploits two core elements: multimodal representations from a contrastive VLM, \ie, CLIP~\cite{clip}, and the information included in large scale vision-language databases (VLD), \eg, PMD~\cite{pmd}. For classification, given an image we retrieve its closest captions in a VLD, encoding both input and captions via the CLIP encoders. We then parse these captions to obtain a set of candidate class names for the input. We encode the candidates via the text encoder of CLIP, scoring them according to their similarity with the visual input and the centroid of the retrieved captions, performing multimodal matching. We name this approach \methodName~(\methodShort).  On a variety of \taskShort\ benchmarks, \methodShort\ achieves higher performance than computationally more expensive VLMs for VQA. We additionally introduce \methodEShort, which extends \methodShort\ with prompt ensembling to further improve the results with minimal computational overhead.

To {further} demonstrate the effectiveness of our proposed approach, we extended \methodShort\ for the task of \taskSName\ (\taskSShort). 
In contrast to the limited nature of image-level classification, often overlooking objects in the background, semantic segmentation aligns with the challenges of unconstrained environments, containing unforeseen objects that cause ambiguities in defining a fixed set of classes. In particular, when faced with the absence of predefined categories, the segmentation task becomes more complex, prompting questions about the appropriate granularity, \eg, whether to segment object parts or the entire object itself. Moreover, segmentation poses new challenges for CaSED, given the tendency of vision-language models to recognize foreground objects and ignore the background, and the object-centric nature of internet-sourced captions.

In this context, we explore different strategies: the first is to use an off-the-shelf segmenter to obtain an initial set of masks. \methodShort\ can then assign a semantic label to each mask independently. A second strategy is to do the opposite: \methodShort\ can provide estimates of the semantic categories that can then be processed by an open-vocabulary segmentation model. Finally, we may avoid any external segmentation model and only employ a single pre-trained VLM, without additional fine-tuning.
Encoding non-overlapping patches separately and classifying their content may seem a good strategy, but we found that it leads to noisy results because a single patch cannot capture the surrounding visual context. To address this issue, we first encode local information of the image by dividing it into cells of different sizes and processing them via CLIP. These multi-scale representations are then accumulated locally to obtain a more precise dense visual representation. The latter undergoes the same \methodShort\ processing, retrieving a set of captions and candidate categories for each local representation. We then apply multimodal scoring on each cell, obtaining the final, dense semantic prediction. We name this approach \methodSShort. Experiments show that \methodSShort\ and \methodShort\ variants outperform various semantic segmentation models in multiple benchmarks, \textit{without} requiring any training procedure.

\noindent To summarize, the contributions of this work are:
\begin{itemize}[noitemsep,nolistsep,leftmargin=*]
    \item We present the tasks of \taskShort\ and \taskSShort, where the goal is to classify/segment images \textit{without} any pre-defined set of target categories, operating directly on an unconstrained semantic space. We define specific metrics for these tasks, capturing the semantic between predictions and ground-truth labels, providing a reference for future research.
    \item We present \methodName, a training-free method for \taskShort\ that exploits multimodal representation of VLMs and captioning database to obtain a coarse set of candidate categories and ranks them according to their multimodal similarity with the input and retrieved captions. We also expand this strategy via prompt ensembling to further improve the performance, naming this variant \methodEShort.
    \item We extend this method for \taskSShort, presenting three variants. While two of them couple \methodShort\ with pretrained segmentation models, the third, \methodSShort\ directly exploits a VLM and multi-scale image processing to obtain local visual representations, that are used to retrieve and score candidates at a local level, providing a dense semantic map of the input without any class priors or training.
    \item Our extensive evaluation on different benchmarks and with different VLM-based models, demonstrate the efficacy of \methodShort\ and \methodEShort\ for \taskShort, and of \methodSShort\ for \taskSShort, highlighting the potential of VLM plus retrieval as a pipeline for semantic categorization tasks with an unconstrained vocabulary.
\end{itemize}

This article extends our previous work~\cite{conti2023vocabulary} in multiple aspects. First, while~\cite{conti2023vocabulary} proposed \taskShort\ and \methodShort, here we extend the latter by exploring improvements of the multimodal scoring mechanism via prompt ensembling.
Moreover, we show the generality of \methodShort\ by applying it to a more recent VLM with a slightly different pre-training objective (\ie, SigLIP~\cite{zhai2023sigmoid}), and including more powerful baselines (\ie, LLaVa~1.5~\cite{liu2023improved}). The main contribution is, however, on the application, as we formalize the task of \taskSName, performing segmentation on an unconstrained vocabulary. In this regard, (i) we propose metrics specific for this task, capturing how well local semantic predictions match with the ground-truth maps, and (ii) we benchmark on multiple datasets (\ie, PascalVOC-20~\cite{pascal_voc_20}, PASCAL Context-59~\cite{pascal_context_59}, and ADE20K-150~\cite{ade20k_150}) and various competitors, even training-based open-vocabulary ones.
We also show three methods to extend 
\methodShort\ for \taskSShort, either using pretrained segmentation models or fully relying on pretrained VLM and captions databases (\methodSShort).
These benchmarks, metrics and results, as well as \methodSShort, will serve as reference for future work aiming to reduce the need of user inputs for semantic segmentation.

\section{Related work}
\label{sec:related}

\noindent\textbf{Vision-Language Models.} The recent surge in models that map image-text pairs into a shared representation space has been largely driven by the availability of large-scale datasets~\cite{laion,schuhmann2022laion,pmd}. These models~\cite{joulin2016learning,gomez2017self,desai2021virtex,clip,jia2021scaling,li2021align,fini2023improved} employ modality-specific encoders and a contrastive objective to align the output representations of the two modalities. A prime example of this approach is CLIP~\cite{clip}, which has demonstrated impressive results in zero-shot classification tasks. Further enhancements to CLIP have been proposed, including the integration of cross-modal attention~\cite{li2021align}, multi-object representation alignment~\cite{zeng2022xvlm}, learning from weak-supervision~\cite{wangsimvlm}, and leveraging unaligned data~\cite{pmd}.
A separate stream of research has focused on enhancing vision-language pre-training for more intricate vision-language tasks, such as image captioning and visual question answering (VQA)~\cite{yu2022coca,hu2022scaling,li2022blip,alayrac2022flamingo,liu2023improved}. Within this domain, BLIP~\cite{li2022blip} uses web data and generated captions to guide the pre-training of a multimodal architecture, surpassing existing VLMs in both captioning and VQA tasks, and LLaVA~\cite{liu2023improved} aligns and CLIP vision encoder and LLAMA large-language model~\cite{touvron2023llama} to reason on visual inputs.

Here, we question a core premise of zero-shot classification with VLMs: the prior knowledge of the target classes. We present \taskShort, that bypasses this assumption, performing classification within an open-ended, language-induced space of semantic categories. In this setting, even advanced methods like BLIP-2~\cite{li2023blip} struggle, whereas caption databases offer valuable priors for deducing the semantic class of an image. It is important to distinguish \taskShort\ from open-vocabulary recognition (\eg, \cite{zareian2021open,ghiasi2022scaling}), as the latter still operates assuming that the list of target classes is known and accessible to the model during inference.

\noindent\textbf{Retrieval augmented models.} In the field of natural language processing, a number of studies have demonstrated the advantages of retrieving information from external databases to enhance the performance of large language models \cite{guu2020retrieval,lewis2020retrieval,borgeaud2022improving}. This approach has also found applications in computer vision, particularly in addressing the issue of class imbalance. For instance, some works have focused on long-tail recognition by learning to retrieve training samples \cite{liu2019large} or image-text pairs from an external database \cite{long2022retrieval}. Another study, \cite{touvron2021grafit}, retrieves images from a specific dataset to learn fine-grained visual representations. More recently, the concept of retrieval-augmentation has been expanded to various types of sources for visual question answering \cite{hu2022reveal}, as well as to condition the generative process in diffusion models \cite{blattmann2022retrieval}, and even image captioning~\cite{ramos2023retrieval}. 

Our work shares similarities with~\cite{long2022retrieval} in that we also utilize an external database. However, unlike~\cite{long2022retrieval} which assumes a pre-defined set of classes (and data) available for training a retrieval module, our approach does not make this assumption due to the vast semantic space of \taskShort. In our method, \methodShort, we use retrieval to first generate a set of candidate classes, and then to perform the final class prediction. Furthermore, we assume the database to contain only captions, and not necessarily paired image-text data, thus making our approach less memory-intensive.

\noindent\textbf{Semantic segmentation.} The earliest works tackling semantic segmentation with an open vocabulary learns a joint embedding space between pixels and class names~\cite{xian2019semantic,bucher2019zero,chen2020uniter,zhao2017open}. After the surge in image-text models such as CLIP~\cite{clip}, the paradigm shifted towards exploiting such web-scale pre-trained models as priors to tackle the problem~\cite{ghiasi2022scaling,liang2023open,xu2023side,yu2023convolutions}.
All prior works on the task assume a fixed pre-defined list of candidate names to address the task of semantic segmentation. Such simplification, however, is unrealistic for real-world applications, where, \eg, a pre-defined list of class names is restrictive and not exhaustive for most scenarios. Different from previous approaches, we aim to tackle a more challenging setup where this list of class names is unavailable and must be inferred from the input.

Recently, we formalized a similar task for image classification~\cite{conti2023vocabulary} and our objective is to expand such scenario to dense classification tasks. The most similar work to our proposed task of \taskSName\ is zero-guidance segmentation~\cite{rewatbowornwong2023zero}. Differently from them, we formalize the \taskSName, strengthening the evaluation protocol, proposing principled metrics for the vocabulary-free scenario, and reusing traditional semantic segmentation benchmarks to assess the performance of multiple baselines methods.

\section{Vocabulary-free Image Classification and Semantic Segmentation}
\label{sec:task}

\noindent\textbf{Preliminaries.} Let us denote as $\mathcal{X}\subset \mathbb{R}^{N \times 3}$ the image space, where $N$ is the number of pixels. Moreover, we can define as $\mathcal{C}$ a set of class labels. These labels are semantic entities in the much larger space of all possible semantic concepts $\mathcal{S}$, \ie, $\mathcal{C}\subset \mathcal{S}$. A classifier/segmenter, is a function $f$ mapping images/pixels to semantic labels in $\mathcal{C}$, with $f:\mathcal{X}\rightarrow \mathcal{C}$ in the case of classification and $f:\mathcal{X}\rightarrow \mathcal{C}^N$ in the case of segmentation. While $f$ is usually trained on paired samples of images and labels, this approach is costly and does not scale with the cardinality of $\mathcal{C}$, as it may require expensive manual annotation. VLMs~\cite{clip,align} removed the need for explicit annotations, measuring similarities between image and text descriptions, \ie, $f_{\mathtt{VLM}}:\mathcal{X} \times \mathcal{T} \rightarrow \mathbb{R}$, with $\mathcal{T}$ the textual space. In this way, we can perform classification by:
\begin{equation}
\label{eq:vlm}
    f(\xvect) = \arg\max_{\cvect \in C} f_\mathtt{VLM}(\xvect,\phi(\cvect))
\end{equation}
where $\phi(\cvect)$ denotes a text concatenation, merging a static text template, or \textit{prompt}, with a class name. 
Segmentation is achieved in a similar manner, computing the similarity between text and local image representations~\cite{liang2023open}.
Note that this does not require any training and the model can classify/segment images into new categories defined at test time without retraining, performing zero-shot transfer. Nevertheless, this approach assumes that the set of categories $\mathcal{C}$ is given a priori by a user. Here, we describe \taskName~(\taskShort) to overcome this limitation in classification, and introduce its counterpart \taskSName~(\taskSShort) for segmentation. 

\noindent\textbf{Task definition.} The goal of the vocabulary-free settings is to either classify (\taskShort) or segment (\taskSShort) an image $\xvect$ without any prior knowledge about $\mathcal{C}$. Specifically, this means operating directly on the vast semantic space $\mathcal{S}$, which encompasses all semantic concepts. For \taskShort, we aim to devise a function $f:\mathcal{X}\rightarrow \mathcal{S}$ that maps an image to a semantic label within $\mathcal{S}$. Similarly, for \taskSShort, the target function is $f:\mathcal{X}\rightarrow \mathcal{S}^N$ mapping images to semantic maps. Note that, at test time, $f$ relies solely on the input image $\xvect$ and a broad repository of semantic concepts approximating $\mathcal{S}$. The task of \taskShort\ is inherently demanding due to the immense number of potential semantic classes in $\mathcal{S}$. For perspective, ImageNet-21k~\cite{deng2009imagenet} is 200 times smaller than the cardinality of the semantic classes in BabelNet~\cite{navigli2010babelnet}. This vast search space presents significant challenges in differentiating nuanced concepts across diverse domains and those with a long-tailed distribution.

\noindent\textbf{Challenges.} 
Both \taskShort~and \taskSShort~share the challenge of identifying which semantic categories in the large set $\mathcal{C}$ are present in the input image~\cite{conti2023vocabulary}. In particular, \taskSShort~is hard as $\mathcal{C}$ contains a lot of potential distractors, \ie, concepts related to the ones in the image but not present. Examples are \textit{couch} vs \textit{sofa}, \textit{tv} vs \textit{monitor}, but also different animal species of plants. Thus, addressing \taskSShort\ requires models with fine-grained recognition capabilities. On the other hand, a model may also segment two regions using synonyms (\eg, \textit{lawn} vs \textit{grass}, \textit{road} vs \textit{highway}): these cases need to be disambiguated to obtain coherent segmentation masks. This also relates to other issues, such as the granularity of the segmentation masks (e.g., parts vs entire objects), or the object-centric focus of VLMs and internet-sourced captions. The latter, tend to ignore objects or elements in the background, making it hard to provide extremely fine-grained segmentation masks.

In the next sections, we describe how we address these challenges by (i) constraining the output space via external captions; (ii) disambiguating semantic concepts via multimodal matching, and (iii) propagating local features.

\section{Method}
\label{sec:ours}

\begin{figure*}[t!]
  \centering
  \includegraphics[width=1\textwidth]{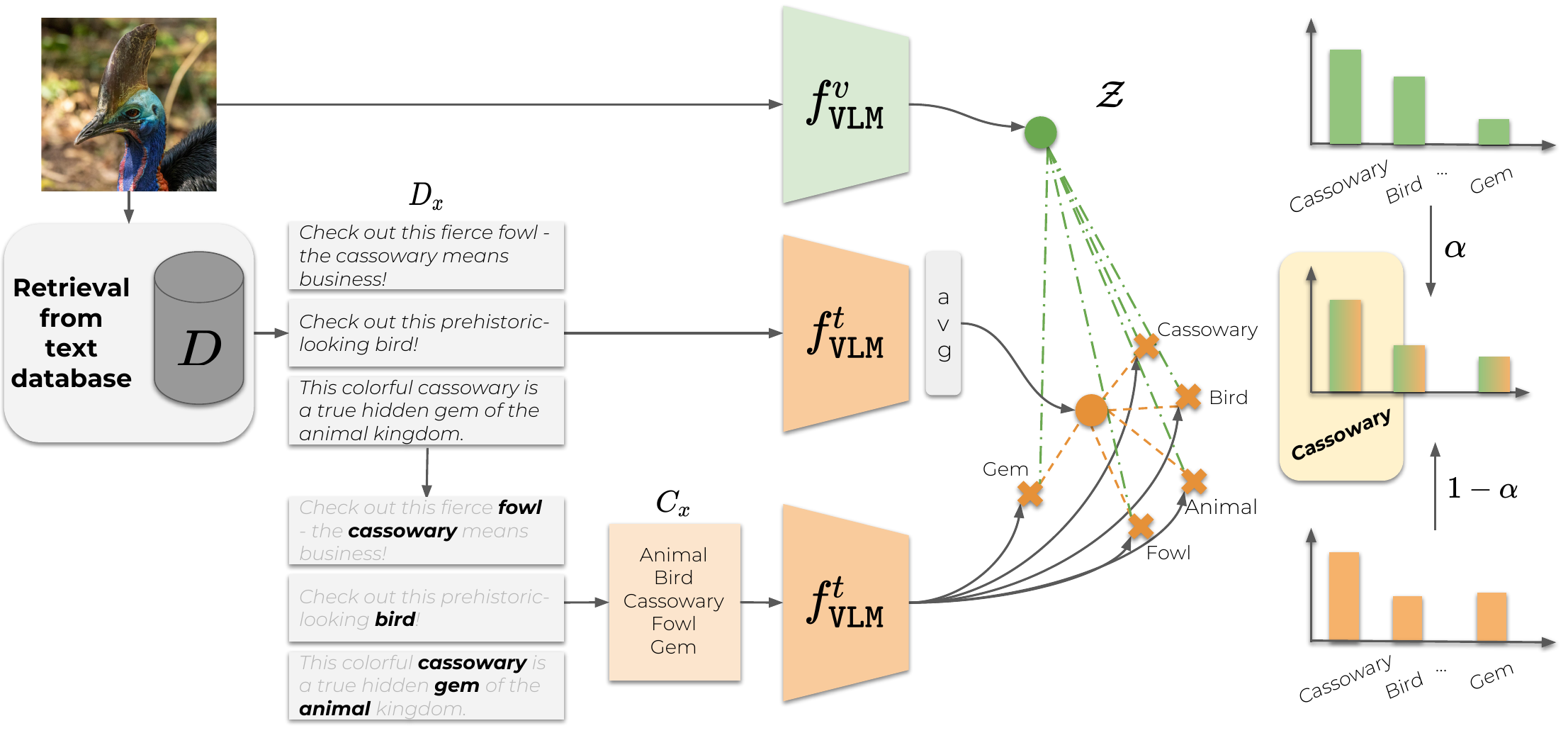}
  \caption{\textbf{\methodShort}.
  Given an input image, \methodShort\ retrieves the most relevant captions from an external database filtering them to extract candidate categories.
  We classify \inlineColorbox{GreenEnrico!20}{image}-to-\inlineColorbox{OrangeEnrico!20}{text} and \inlineColorbox{OrangeEnrico!20}{text}-to-\inlineColorbox{OrangeEnrico!20}{text}, using the retrieved captions centroid as the textual counterpart of the input image.}
  \label{fig:neurips}

\end{figure*}

In the following, we first describe \methodName\ (\methodShort)~\cite{conti2023vocabulary}, for tackling \taskShort. The method leverages the power of large Vision-Language Datasets (VLDs) to find the best matching category within an unconstrained semantic space. We then describe some modifications that allow us to improve performance of the method while also increasing its speed.
Finally, we present how \methodShort\ can be extended to semantic segmentation, either by application on top of pretrained segmentation network or by modifying how CLIP processes the input image  (\methodSShort).
Fig.~\ref{fig:neurips} shows an overview of \methodShort.

\subsection{\methodName}

\methodShort\ is built on two modules: (i) candidate categories generation an (ii) multimodal scoring of the list of candidates. In the following, we describe each of them, and how we further improve the performance and speed in \methodEShort.

\noindent\textbf{Candidate category generation.}
We initially narrow down the vast classification space $\mathcal{S}$ to few probable candidate classes. Given an input $\xvect$, we use the pre-trained VLM $f_\mathtt{VLM}$ and an external image captions database $D$ to retrieve a subset $D_{\xvect} \subset D$ of $K$ closest captions to the input image as
\begin{equation}
\label{eq:retrieval}
    D_{\xvect} = \topk_{\dvect\in D} \, f_{\mathtt{VLM}} (\xvect,\dvect) = \topk_{\dvect\in D} \;\langle f^{v}_\mathtt{VLM}(\xvect), f^t_\mathtt{VLM}(\dvect)\rangle,
\end{equation}
where $f^v_\mathtt{VLM}:\mathcal{X}\rightarrow \mathcal{Z}$ and $f^t_\mathtt{VLM}:\mathcal{T}\rightarrow \mathcal{Z}$ are the visual and textual encoders of the VLM, respectively, and $\mathcal{Z}$ is their shared embedding space. The operation $\langle\cdot,\cdot \rangle$ computes the cosine similarity between the two. Our approach can accommodate varying database sizes and is not dependent on the specific form of $D$. We then extract a finite set of candidate classes $C_{\xvect}$ from $D_{\xvect}$ using basic text parsing and filtering techniques. More details in Appendix~\ref{sec:candidates}.

\noindent\textbf{Multimodal candidate scoring.} 
We score each candidate in the set $C_{\xvect}$ using both visual and textual semantic similarities via the VLM encoders to identify the best-matching class for the input image. We denote $s^v_{\cvect}$ as the visual score of each candidate category $\cvect$, computed as the similarity between the visual representation of the input image and the textual representation of the candidate name:
\begin{equation}
    \label{eq:visual_score}
    s^v_{\cvect} = \langle f^v_\mathtt{VLM}({\xvect}), f^t_\mathtt{VLM}({\cvect}) \rangle.
\end{equation}
To mitigate the modality gap in the space $\mathcal{Z}$, we introduce a unimodal text-to-text scoring. Denoting the centroid $\bar{\dvect_{\xvect}}$ of the retrieved captions, the text-based matching score $s^t_{\cvect}$ is:
\begin{equation}
    \label{eq:text_mean}
    \bar{\dvect_{\xvect}} = \frac{1}{K}\sum_{\dvect\in D_{\xvect}} f^t_\mathtt{VLM}({\dvect}),
\end{equation}
\begin{equation}
    \label{eq:text_score}
    s^t_{\cvect} = \langle \bar{\dvect_{\xvect}}, f^t_\mathtt{VLM}(\cvect) \rangle.
\end{equation}
The use of the caption centroids as anchor for classification is motivated by the analyses in \cite{conti2023vocabulary}, showing how they are well correlated with the semantic of the visual input. We report this analysis in Appendix~\ref{sec:semantic_space_rep}. We obtain the final score $s_{\cvect}$ for each candidate $\cvect$ by merging the two scores, as:
\begin{equation}
    \label{eq:final_score}
    s_{\cvect} =  \alpha\;\sigma({s}^v_{\cvect}) \;+ \;(1-\alpha)\; \sigma(s^t_{\cvect})
\end{equation}
where $\sigma(\cdot)$ is the softmax operation on the two scores of each candidate class, and $\alpha$ is a hyperparameter. The output category is $f_\mathtt{CaSED}(\xvect) = \arg\max_{{\cvect}\in C_{\xvect}} s_{\cvect}$. Our approach, \methodShort, is \textit{training-free}, uses a pre-trained and frozen VLM, and is flexible for various architectures and databases.

\subsection{\methodEShort}

\begin{figure*}[t!]
\centering
\begin{tabular}{ccc}
\includegraphics[width=0.3\linewidth]{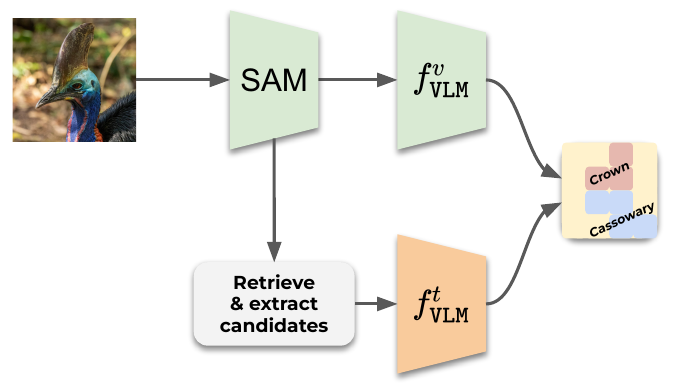} & \includegraphics[width=0.3\linewidth]{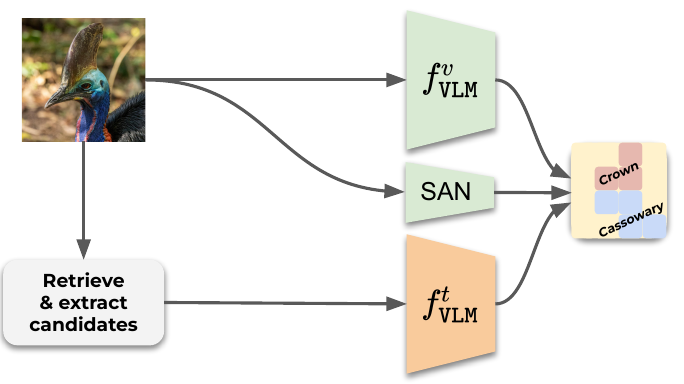} & \includegraphics[width=0.3\linewidth]{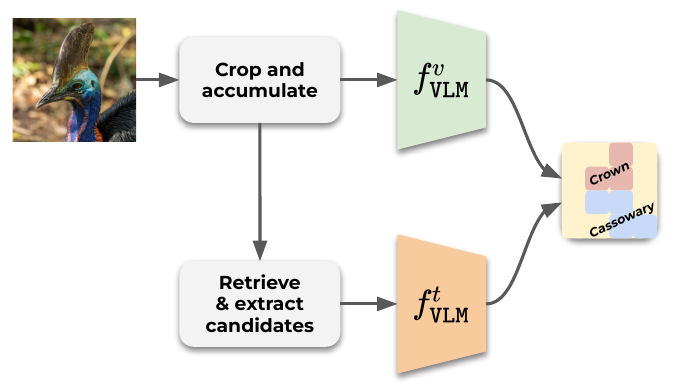} \\
(a) SAM-based method & (b) SAN-based method & (c) \methodSShort\ \\
\end{tabular}
\caption{\textbf{Extending CaSED for Semantic Segmentation}. We follow three strategies: (a) a class-agnostic segmenter (SAM) segments all objects, then CaSED labels each mask independently; (b)  CaSED provides candidate categories for the image that are fed as input to an open-vocabulary segmentation model (SAN); (c) DenseCaSED, where we directly accumulate visual features from multi-scale patches, and perform CaSED locally. }
 \label{fig:semantic_seg_methods}
\end{figure*}

To improve the performance of \methodShort, we propose a simple yet effective modification, introducing prompt ensembling~\cite{clip} after the caption extraction and filtering.
Prompt ensembling applies a predefined list of templates to the class names to enhance their contextual information. For instance, the class name "dog" could be expanded to "a photo of a dog". By applying a set of templates rather than a single one and computing their average representation, the resulting features better capture the semantic meaning of the word, leading to consistent performance gains~\cite{clip}.

We apply prompt ensembling to the candidates generated from the retrieved captions. The number of templates used is variable, depending on the specific dataset\footnote{For each classification dataset, we use the templates defined in CLIP~\cite{clip}, spanning from 1 for, \eg, Flowers-102~\cite{flowers102}, Food-101~\cite{food101}, and Oxford Pets, to 48 for UCF101~\cite{ucf101}}. The average representation of these ensembled prompts is then used as anchors to compute the image-to-text and text-to-text scores as described above. Formally, for a candidate $\cvect$ and a set of templates $T$, the ensembled representation $f^t_{\text{ens}}(\cvect)$ is computed as:
\begin{equation}
    f^t_{\text{ens}}(\cvect) = \frac{1}{|T|}\sum_{t\in T} f^t_\mathtt{VLM}(t(\cvect)),
\end{equation}
where $t(\cvect)$ means applying template $t$ to candidate $\cvect$. $f^t_{\text{ens}}(\cvect)$ is then used for image-to-text and text-to-text scoring.

\section{\methodShort\ for Vocabulary-free Semantic Segmentation}

To extend \methodShort\ for the task of semantic segmentation, we can follow three strategies. The first is to exploit an available class-agnostic segmentation model~\cite{kirillov2023segment}, extract segmentation masks and then assigning them a label via \methodShort, independently. The second does the opposite: the initial set of candidates generated by \methodShort\ is the input to an open-vocabulary segmentation network~\cite{xu2023side}. While these strategies lead to good results but they require the additional computational cost of using another segmentation network. We thus propose a third approach which generates local visual representations directly from patches of the input image. In the following, we describe the three strategies.

\subsection{Coupling \methodShort\ with Segmentation Networks}
\noindent{\textbf{{Assigning semantics to class-agnostic masks with \methodShort}}.} Let us assume to have a class-agnostic segmentation network ${f}_{\mathtt{SEG}}$ that, given as input an image $\xvect$ maps it to a set of $k$ segmentation masks ${M} = \{\mvect_1, \cdots, \mvect_k\}$. Note that these masks have no semantic attached, and the number of masks $k$ may be input dependent~\cite{kirillov2023segment}.
From the masks, we extract a set of image regions ${R} = \{\rvect_1, \cdots, \rvect_k\}$, \eg, by cropping around the relative mask. We can then attach a semantic to each region using \methodShort, propagating the prediction to the pixels of the relative mask. Given a pixel $x \in \xvect$ that assigned to mask $\mvect_i \in M$ by ${f}_{\mathtt{SEG}}$, its semantic label is simply $f_\mathtt{CaSED}(\rvect_i)$. The rationale behind this approach is that the semantic of a pixel is the one assigned by \methodShort\ to an image extracted from the mask the pixels belong to. 

\noindent{\textbf{Candidates generation with \methodShort\ for Open-Vocabulary Segmentation}}.
The main limitation of the previous approach is that extracting meaningful image regions from masks may require solutions that exploit VLM priors on object localization (\eg, circle drawing~\cite{shtedritski2023does}). To sidestep this problem, we can invert the pipeline by first obtaining candidate class-names using \methodShort\ and then segmenting the image using an open-vocabulary segmentation model, \eg, \cite{san}. Specifically, an open-vocabulary segmentation model takes as input an image and a set of possible labels $\mathcal{Y}\subseteq \mathcal{S}$ and maps them to a segmentation mask, assigning pixels to elements of $\mathcal{Y}$, \ie, $f_\mathtt{OV-SEG}:\mathcal{X}\times\mathcal{Y}\rightarrow\mathcal{Y}^N$. As we will show experimentally in Sec.~\ref{sec:experiments}, if we do not take into account the challenges of \taskSShort\ and set $\mathcal{Y}=\mathcal{S}$, we obtain poor segmentation results. This is mainly due to the extremely large cardinality of $\mathcal{S}$, requiring to distinguish local, fine-grained differences of potentially similar semantic concepts. 

To overcome these challenges, we follow the rationale behind \methodShort, restricting the search space by estimating a set of candidate classes. Thus, given an image $\xvect$, we define a set of candidates $C_{\xvect}$ by filtering a set of captions $D_{\xvect}$, with the latter obtained as in Eq.~\eqref{eq:retrieval}. We then feed the set $C_{\xvect}$ as input to the open-vocabulary segmentation network and obtain the relative segmentation mask as $f_\mathtt{OV-SEG}(\xvect,C_{\xvect})$.

\subsection{\methodSShort}
The previous approaches rely on the presence of an additional (pretrained) semantic segmentation model, and the assumption that the module is not biased toward particular input distributions. Here we propose a different strategy, directly exploiting the available VLM.  

The approach applies \methodShort\ to local image representations, as done with the class-agnostic strategy previously described. As we do not have access to masks, we need to define local image regions that we can feed as input to the VLM. To obtain such representations, 
We divide the image in multiple grids, where each grid has $n^2$. For simplicity, we choose $n$ to be powers of 2, \ie, $n\in \{1, 2, 3\}$. We also replicate the grids by shifting them vertically and/or horizontally by a stride equal to half the size of a grid cell.
This creates a hierarchy of patches where neighboring patches are likely to belong to the same super patch and therefore their representation (loosely) depends on each other. This helps in embedding contextual information in the aggregated pixel-level representation.%}. 

Formally, let us denote the visual representation of a pixel $i$ in $\xvect$ as $\lvect_{i}$. Moreover, let us denote as $\{\gvect^1_{i}, \cdots, \gvect^{N}_{i}\}$ all patches that contain $i$.
The local representation $\lvect_{i,j}$ is then
\begin{equation}
    \label{eq:local-representation}
    \lvect_{i} = \frac{1}{N} \sum_{q=1}^{N}
    f^{v}_\mathtt{VLM}(\gvect^q_{i}).
\end{equation}
 Note that we divide the aggregated value for each pixel by the number of times it was forwarded within a cell.
As these local representations are already encoded using a VLM, we can retrieve the most relevant set of captions to a cell from a VLD $D$ using the cosine similarity of the embeddings, as:
\begin{equation}
\label{eq:retrieval-local}
    D_{{\lvect}_{i}} = \topk_{\dvect\in D} \;\langle {\lvect}_{i}, f^t_\mathtt{VLM}(\dvect)\rangle,
\end{equation}
For each cell in $\gvect\in G_g$, we follow the pipeline of \methodShort, by (i)  filtering $D_{\hat{\lvect}}$ to obtain the corresponding set of candidates $\mathcal{C}_{\gvect}$; (ii) computing the visual score as in  Eq.~\eqref{eq:visual_score} but using $\hat{\lvect}$ as visual representation; (iii) compute the textual scoring as in Eq.~\eqref{eq:text_score} with $D_{\hat{\lvect}}$ as set of captions; (iv) merging the two scores to compute the final multimodal one, as in Eq.~\eqref{eq:final_score}. The \methodShort\ predictions on accumulated local visual features are then propagated to the whole cell, producing the final segmentation mask. We name this approach \methodSShort.

Note that this approach is not only training-free but does not use any segmentation network, relying only on a contrastive-based VLM. Moreover, accumulating local cells representations across scales allows to model the context in which a cell appears while enforcing a consistent vocabulary across neighboring cells. This latter aspect is important in \taskSShort, as modeling cells in isolation may lead to inconsistent choices of labels in the large search space (e.g. "sofa" vs "couch") leading to lower segmentation results.

\section{Experiments}
\label{sec:experiments}

\begin{table*}[ht]
\centering
\begin{tabular}{cc|cccccccccc|c}
\toprule
\multicolumn{2}{c|}{\textbf{Method}}  & \multicolumn{11}{c}{\textbf{Cluster Accuracy (\%)} $\uparrow$} \\
&& {C101} & {DTD} & {ESAT} & Airc. & Flwr & Food & {Pets} & {SUN} & {Cars} & {UCF} & \textbf{Avg.} \\
\midrule
\scriptsize\parbox[t]{2mm}{\multirow{2}{*}{\rotatebox{90}{CLIP}}}
 & WordNet & 34.0 & 20.1 & 16.7 & 16.7 & 58.3 & 40.9 & 52.0 & 29.4 & 18.6 & 39.5 & 32.6\\
 & English Words & 29.1 & 19.6 & 22.1 & 15.9 & 64.0 & 30.9 & 44.4 & 24.2 & 19.3 & 34.5 & 30.4 \\
 \midrule
\scriptsize\parbox[t]{2mm}{\multirow{4}{*}{\rotatebox{90}{Caption}}} & Closest Caption & 12.8 & 8.9 & 16.7 & 13.3 & 28.5 & 13.1 & 15.0 & 8.6 & 20.0 & 17.8 & 15.5 \\
 & BLIP-2 (ViT-L)& 26.5 & 11.7 & 23.3 & 5.4 & 23.6 & 12.4 & 11.6 & 19.5 & 14.8 & 25.7 & 17.4 \\
 & BLIP-2 (ViT-g) & 37.4 & 13.0 & 25.2 & 10.0 & 29.5 & 19.9 & 15.5 & 21.5 & 27.9 & 32.7 & 23.3 \\
 & LLaVA 1.5 (7B) & 41.1 & 11.1 & 19.7 & 10.4 & 13.4 & 11.1 & 12.8 & 14.0 & 12.1 & 29.5 & 17.5 \\
 \midrule
\scriptsize\parbox[t]{2mm}{\multirow{3}{*}{\rotatebox{90}{VQA}}} & {BLIP-2 (ViT-L)} & 60.4 & 20.4 & 21.4 & 8.1 & 36.7 & 21.3 & 14.0 & 32.6 & 28.8 & 44.3 & 28.8 \\
 & {BLIP-2 (ViT-g)} & 62.2 & 23.8 & 22.0 & 15.9 & 57.8 & 33.4 & 23.4 & 36.4 & \textbf{57.2} & \textbf{55.4} & 38.7 \\
 & LLaVA 1.5 (7B) & \textbf{76.2} & \textbf{30.6} & \textbf{38.9} & 3.0 & 5.8 & 22.7 & 7.7 & 27.5 & 2.6 & 48.0 & 26.3 \\
 \midrule
 \rowcolor{ForestGreen!20} \multicolumn{2}{c|}{\methodShort} & 51.5 & 29.1 & 23.8 & 22.8 & 68.7 & 58.8 & 60.4 & 37.4 & 31.3 & 47.7 & 43.1 \\
 \rowcolor{ForestGreen!20} \multicolumn{2}{c|}{\methodEShort} & 51.3 & 29.3 & 21.0 & \textbf{24.4} & \textbf{70.4} & \textbf{61.2} & \textbf{60.9} & \textbf{37.7} & 38.5 & 46.6 & \textbf{44.1} \\
 \midrule
\rowcolor{Gray!20}\multicolumn{2}{c|}{CLIP upper bound}& 87.6 & 52.9 & 47.4 & 31.8 & 78.0 & 89.9 & 88.0 & 65.3 & 76.5 & 72.5 & 69.0 \\
\bottomrule
\end{tabular}
\vspace{1mm}
\caption{Cluster Accuracy on the ten datasets. \inlineColorbox{ForestGreen!20}{Green} is our method, \inlineColorbox{Gray!20}{gray} shows the upper bound.}
\label{tab:metric_semantic_cluster_acc}
\end{table*}

\begin{table*}[t]
\centering
\begin{tabular}{cc|cccccccccc|c}
\toprule
\multicolumn{2}{c|}{\textbf{Method}}  &\multicolumn{11}{c}{\textbf{Semantic IoU (\%) $\uparrow$}} \\
&& {C101} & {DTD} & {ESAT} & Airc. & Flwr & Food & {Pets} & {SUN} & {Cars} & {UCF} & \textbf{Avg.} \\
\midrule
 \scriptsize\parbox[t]{2mm}{\multirow{2}{*}{\rotatebox{90}{CLIP}}}
 & WordNet & 15.0 & 3.0 & 1.3 & 0.5 & 31.3 & 7.8 & 14.7 & 9.0 & 4.8 & 3.8 & 9.1 \\
 &  English Words & 8.0 & 2.0 & 0.0 & 1.1 & 16.4 & 2.0 & 17.2 & 8.1 & 2.7 & 1.8 & 5.9 \\\midrule
 \scriptsize\parbox[t]{2mm}{\multirow{4}{*}{\rotatebox{90}{Caption}}}
 & Closest Caption & 4.5 & 0.8 & 1.3 & 1.9 & 5.9 & 3.1 & 3.0 & 2.3 & 11.4 & 1.0 & 3.5 \\
 & BLIP-2 (ViT-L) & 13.4 & 1.4 & 4.8 & 0.0 & 7.5 & 4.7 & 1.7 & 4.7 & 11.6 & 1.1 & 5.1 \\
 & BLIP-2 (ViT-g) & 16.8 & 1.8 & 4.1 & 0.1 & 13.9 & 7.9 & 2.9 & 5.7 & 24.7 & 1.9 & 8.0 \\
 & LLaVA 1.5 (7B) & 13.8 & 0.87 & 5.13 & 0.0 & 2.9 & 1.7 & 0.3 & 3.5 & 4.6 & 1.4 & 3.4 \\
 \midrule
\scriptsize\parbox[t]{2mm}{\multirow{3}{*}{\rotatebox{90}{VQA}}}
 & {BLIP-2 (ViT-L)} & 36.1 & 1.8 & 7.0 & 0.1 & 21.5 & 3.7 & 5.7 & 11.5 & 18.9 & 2.5 & 10.9 \\
 & {BLIP-2 (ViT-g)} & \textbf{41.5} & 2.4 & \textbf{7.5} & 2.0 & \textbf{38.0} & 8.6 & 10.2 & 13.8 & \textbf{33.2} & 2.8 & 16.0 \\
 & LLaVA 1.5 (7B) & 41.4 & 0.4 & 6.1 & 0.0 & 5.0 & 3.0 & 0.7 & 6.6 & 1.3 & 2.0 & 6.6 \\
 \midrule
 \rowcolor{ForestGreen!20} \multicolumn{2}{c|}{\methodShort} & 35.4 & \textbf{5.1} & 2.3 & \textbf{4.8} & 33.1 & \textbf{19.4} & \textbf{35.1} & 17.2 & 16.2 & \textbf{8.4} & 17.7 \\
 \rowcolor{ForestGreen!20} \multicolumn{2}{c|}{\methodEShort} & 37.8 & 4.6 & 5.2 & 4.4 & 35.2 & 18.9 & 34.9 & \textbf{17.8} & 16.0 & 7.8 & \textbf{18.3} \\
 \midrule
  \rowcolor{Gray!20} \multicolumn{2}{c|}{CLIP upper bound} & 86.0 & 52.2 & 51.5 & 28.6 & 75.7 & 89.9 & 88.0 & 66.6 & 84.5 & 71.3 & 69.4 \\
\bottomrule
\end{tabular}
\vspace{1mm}
\caption{Semantic IoU on the ten datasets. \inlineColorbox{ForestGreen!20}{Green} is our method, \inlineColorbox{Gray!20}{gray} shows the upper bound.}
\label{tab:metric_semantic_iou}
\end{table*}

\begin{table*}[t]
\centering
\begin{tabular}{cc|cccccccccc|c}
\toprule
\multicolumn{2}{c|}{\textbf{Method}}  &\multicolumn{11}{c}{\textbf{Semantic Similarity (x100)} $\uparrow$} \\
&& {C101} & {DTD} & {ESAT} & Airc. & Flwr & Food & {Pets} & {SUN} & {Cars} & {UCF} & \textbf{Avg.} \\
\midrule
\scriptsize\parbox[t]{2mm}{\multirow{2}{*}{\rotatebox{90}{CLIP}}}
 & WordNet & 48.6 & 32.7 & 24.4 & 18.9 & 55.9 & 49.6 & 53.7 & 44.9 & 28.8 & 44.2 & 40.2 \\
 &  English Words & 39.3 & 31.6 & 19.1 & 18.6 & 43.4 & 38.0 & 44.2 & 36.0 & 19.9 & 34.7 & 32.5 \\ \midrule
\scriptsize\parbox[t]{2mm}{\multirow{4}{*}{\rotatebox{90}{Caption}}}
 & Closest Caption & 42.1 & 23.9 & 23.4 & 29.2 & 40.0 & 46.9 & 40.2 & 39.8 & 49.2 & 40.3 & 37.5 \\
 & BLIP-2 (ViT-L) & 57.8 & 31.4 & 39.9 & 24.4 & 36.1 & 44.6 & 29.0 & 45.3 & 46.4 & 38.0 & 39.3 \\
 & BLIP-2 (ViT-g) & 63.0 & 33.1 & 36.2 & 24.3 & 45.2 & 51.6 & 31.6 & 48.3 & 61.0 & 44.6 & 43.9 \\
 & LLaVA 1.5 (7B) & 56.8 & 29.4 & 40.5 & 21.3 & 31.1 & 36.9 & 24.8 & 42.5 & 38.1 & 37.9 & 35.9 \\
 \midrule
\scriptsize\parbox[t]{2mm}{\multirow{3}{*}{\rotatebox{90}{VQA}}}
 &{BLIP-2 (ViT-L)} & 70.5 & 34.9 & 29.7 & 29.1 & 48.8 & 42.0 & 40.0 & 50.6 & 52.4 & 48.6 & 44.7 \\
 & {BLIP-2 (ViT-g)} & \textbf{73.5} & 36.5 & 31.4 & \textbf{30.8} & \textbf{59.9} & 52.1 & 43.9 & \textbf{53.3} & \textbf{65.1} & \textbf{55.1} & 50.1 \\
 & LLaVA 1.5 (7B) & 72.6 & 36.7 & \textbf{44.1} & 29.4 & 41.8 & 41.1 & 36.0 & 41.9 & 35.3 & 46.6 & 42.6 \\
 \midrule
\rowcolor{ForestGreen!20} \multicolumn{2}{c|}{\methodShort} & 65.7 & 40.0 & 32.0 & 30.3 & 55.5 & 64.5 & 62.5 & 52.5 & 47.4 & 54.1 & 50.4 \\
\rowcolor{ForestGreen!20} \multicolumn{2}{c|}{\methodEShort} & 66.3 & \textbf{40.3} & 34.9 & 27.0 & 56.1 & \textbf{65.0} & \textbf{62.9} & 53.0 & 46.5 & 53.8 & \textbf{50.6} \\
\midrule
 \rowcolor{Gray!20}\multicolumn{2}{c|}{CLIP upper bound} & 90.8 & 69.8 & 67.7 & 66.7 & 83.4 & 93.7 & 91.8 & 80.5 & 92.3 & 83.3 & 82.0 \\
\bottomrule
\end{tabular}
\vspace{1mm}
\caption{Semantic Similarity (x100) on the ten datasets. Values multiplied by x100 for readability. \inlineColorbox{ForestGreen!20}{Green} highlights our method and \inlineColorbox{Gray!20}{gray} the upper bound.}
\label{tab:metric_semantic_similarity}
\end{table*}

\subsection{Classification}

\noindent\textbf{Datasets.}
As in previous studies~\cite{shu2022test,zhou2022learning}, we use ten datasets that span both broad and detailed classification in various domains. These datasets include Caltech-101~(C101)~\cite{caltech101}, DTD~\cite{dtd}, EuroSAT~(ESAT)~\cite{eurosat}, FGVC-Aircraft~(Airc.)~\cite{fgvc_aircraft}, Flowers-102~(Flwr)~\cite{flowers102}, Food-101~(Food)~\cite{food101}, Oxford Pets~(Pets), Stanford Cars~(Cars)~\cite{stanford_cars}, SUN397~(SUN)~\cite{sun397}, and UCF101~(UCF)~\cite{ucf101}. For tuning hyperparameters, we use the ImageNet dataset~\cite{deng2009imagenet}.

\noindent\textbf{Evaluation metrics.}
The unrestricted nature of the semantic space in \taskShort\ requires unique evaluation metrics. In~\cite{conti2023vocabulary}, we propose two primary measures: \textit{semantic relevance}, which assesses the similarity between the predicted and actual labels, and \textit{image grouping}, which evaluates the quality of image clustering based on the predicted labels. For semantic relevance, we consider two aspects: i) \textit{Semantic Similarity}, which measures the similarity between the predicted and actual labels in a semantic space, and ii) \textit{Semantic Intersection over Union (IoU)}, which calculates the overlap of words between the prediction and the ground truth. Formally, given an input $\xvect$ with ground-truth label $\yvect$ and prediction $\hat{\cvect} = f(\xvect)$, the \textit{Semantic Similarity} is computed as $\langle g(\hat{\cvect}), g(\yvect) \rangle$, where $g: \mathcal{T} \rightarrow \mathcal{Y}$ is a function that maps text to an embedding space $\mathcal{Y}$. To accommodate free-form text, we employ Sentence-BERT~\cite{reimers2019sentence} as $g$. For \textit{Semantic IoU}, given a predicted label $\cvect$ (considered as a set of words), we compute the Semantic IoU as $|\cvect \cap \yvect|/|\cvect \cup \yvect|$, where $\yvect$ is the set of words in the ground-truth label. To evaluate image grouping, we use the traditional \textit{Cluster Accuracy} metric inspired by protocols for deep visual clustering~\cite{van2020scan,ji2019invariant,han2019automatically}. This involves clustering images based on their predicted labels and then assigning each cluster to a ground-truth label with a many-to-one match, where a predicted cluster is assigned to the most common ground-truth label.

\noindent\textbf{Baselines.} We consider a diverse set of baselines, categorized into three groups. The first uses CLIP with extensive vocabularies, such as WordNet~\cite{wordnet}, which contains approximately 117k names, and the English Words dataset~\cite{web2}, featuring around 234k names. As a theoretical upper limit, we evaluate CLIP with an ideal vocabulary, specifically the ground-truth names from the target dataset (CLIP upper bound). While we primarily present results for CLIP with the ViT-L architecture~\cite{dosovitskiyimage} due to space constraints, additional results utilizing other architectures are in Appendix~\ref{sec:ablation_backbone}. The second group of baselines includes captioning methods, directly describing the semantic content of images. We explore two approaches: one that retrieves captions from a database and another that generates captions using a pre-trained image captioning model. For caption retrieval, we use the same VLD of \methodShort.
For caption generation, we use BLIP-2~\cite{li2023blip}, a VLM known for its exceptional performance across various tasks, including image captioning, to generate image descriptions. The third group use a Visual Question Answering (VQA) model to directly infer the class in the image. We use BLIP-2~\cite{li2023blip}
, and extend the baselines of~\cite{conti2023vocabulary} with 
LLaVA 1.5 (7B)~\cite{liu2023improved}, a larger VLM.

\noindent\textbf{Implementation Details.} We conduct our experiments on NVIDIA A6000 GPUs, using mixed-precision for efficiency. We use a subset of PMD~\cite{pmd} as our database, which includes five of its largest datasets: Conceptual Captions (CC3M)~\cite{cc3m}, Conceptual Captions 12M (CC12M)~\cite{cc12m}, Wikipedia Image Text (WIT)~\cite{wit}, Redcaps~\cite{redcaps}, and the portion of YFCC100M*~\cite{yfcc100m} curated for PMD. For retrieval, we embed the database with the CLIP text encoder $f^t_\mathtt{VLM}$, using a fast indexing technique, \ie, FAISS~\cite{faiss}. The hyperparameter $\alpha$ in Eq.~\eqref{eq:final_score} and the number of retrieved captions $K$ are fixed to $\alpha=0.7$ and $K=10$ via selection on ImageNet.

\noindent\textbf{Quantitative results.} The performance of \methodShort\ is superior to all baseline models across all metrics, as shown in Tab.~\ref{tab:metric_semantic_cluster_acc}, Tab.~\ref{tab:metric_semantic_similarity}, and Tab.~\ref{tab:metric_semantic_iou}. Notably, \methodShort\ outperforms BLIP-2 (VQA) over ViT-g by $+4.4\%$ in Cluster Accuracy and $+1.7\%$ on Semantic IoU, while using significantly fewer parameters. The performance gap is even wider when compared to the same visual backbone, with \methodShort\ surpassing BLIP-2 on ViT-L (VQA) by $+14.3\%$ on Cluster Accuracy, $+5.7$ in Semantic Similarity, and $+6.8\%$ on Semantic IoU. In addition, the CLIP with retrieval a large pre-defined vocabulary does not yield effective results for \taskShort, likely due to the challenge of classifying over a vast search space with difficult-to-model class boundaries. This is evidenced by CLIP with English Words and with WordNet underperforming across all metrics. Captioning models, despite their ability to capture image semantics in challenging settings, exhibit high variability across images of the same category, resulting in poor performance on Clustering and Semantic IoU. LLaVA 1.5 (VQA), while generally underperforming compared to BLIP-2 (VQA) and \methodShort, does achieve the best results for cluster accuracy on Caltech-101, DTD, and EuroSAT. However, its performance on semantic IoU and semantic similarity is not significantly better than other approaches, and on average, it achieves unsatisfactory results across the ten datasets. These results highlight the efficacy of \methodShort\ in all metrics. Finally, \methodEShort\ consistently improves performance across all datasets, with an average gain of $+1.0\%$, $+0.6$, and $+0.2$ on the three metrics w.r.t. \methodShort.

\subsection{Semantic segmentation}

\noindent\textbf{Datasets.}
We experiment with three datasets: Pascal VOC~\cite{pascal_voc_20} (VOC-20), PASCAL Context-59~\cite{pascal_context_59} (CTX-59), and ADE20K-150~\cite{ade20k_150} (ADE-150). As opposed to~\cite{san,odise,ovseg}, we do not use COCO Stuff~\cite{coco_stuff}, as it is used for training and all the considered baselines and methods are training-free.

\noindent\textbf{Evaluation metrics.}
To address the openness \taskSShort, we extend two popular metrics for open-vocabulary semantic segmentation, namely Jaccard Index (JI) and Recall (R). We will refer to the metrics as Hard JI (HJI) and Hard Recall (HR). For their soft variants, we replace the binary "hard" values (\ie, zero and one) with the semantic similarity between predicted and ground-truth word. The Soft Jaccard Index (SJI), directly accounts for this similarity at pixel-level. Formally, for a pixel $\pvect$ with label $\yvect$ and prediction $\hat{\cvect} = f(\pvect)$, the textual similarity is computed as $\langle g(\hat{\cvect}), g(\yvect) \rangle$, where the function $g: \mathcal{T} \rightarrow \mathcal{Y}$ maps text to an embedding space $\mathcal{Y}$. As in~\cite{conti2023vocabulary}, we use Sentence-BERT~\cite{reimers2019sentence} as $g$. Similarly, Soft Recall (SR) expands the recall metric with the semantic proximity of the predicted and ground-truth classes in the image.

Furthermore, we introduce two variants of the JI, \ie, Nearest Jaccard Index~(NJI) and Overlap Jaccard Index~(OJI) to account for cases where proposed words may not perfectly align with the annotations due to linguistic ambiguities or to specificity of the proposed segmentation masks, \ie, part vs whole cases (\eg, predicting "head" and "shirt" on two parts of a "person"). We propose to map predicted names to ground-truth ones, evaluating the traditional JI on the projected predicted mask. More formally, given the predicted segmentation mask $\Cvect$ and its ground-truth mask $\Yvect$, we extract the lists of predicted names $L^{\Cvect} \in \Cvect$ and the ground-truth names $L^{\Yvect} \in \Yvect$. We then create a mapping $M: \mathcal{L^{\Cvect}} \rightarrow \mathcal{L^{\Yvect}}$, so that each predicted word is mapped to one ground-truth word. The criteria behind this mapping is the main difference between NJI and OJI. In the former, we use textual similarity between predictions and the list of annotated words.
In the latter, we directly evaluate the co-occurrence of predictions and annotations in the pixel space.

\noindent\textbf{Baselines.} We consider two groups of baseline methods for comparison. The first exploits SAM~\cite{kirillov2023segment} to first extract regions and then propose region-based candidate names via retrieval or generation with, \eg, English Words or BLIP-2. The second uses a open-vocabulary semantic segmentation model in the absence of the pre-defined list of class names for the dataset, requiring the ad-hoc generation with traditional vocabulary-free methods. In this context, we employ SAN~\cite{san} as it offers a minor variation from the CLIP architecture, incorporating only an auxiliary network to address the task on the pre-trained backbone. For both groups, we report results with the same set of baselines of \taskShort, without the captioning baselines. We also report results when \methodShort\ is used for vocabulary generation. For all main experiments, we use CLIP with the ViT-L/14 backbone.

\noindent\textbf{Implementation details.}
We conduct all our experiments following the same setup as for classification. \methodSShort\ introduces a single hyperparameter, \ie, the grid sizes to crop the input image. We empirically set $N=\{2, 4, 8\}$ to have a final dense pixel map of $16\times16$, for a total of 256 sub-regions for each input image. For computational efficiency, we use FastSAM~\cite{zhao2023fast} in place of SAM for segmentation.

\begin{table}[t]
\centering
\begin{tabular}{cc|cccc|cc}
\toprule
\multicolumn{2}{c|}{\textbf{Method}}  & \multicolumn{6}{c}{\textbf{VOC-20}} \\
& & \multicolumn{4}{c|}{\textbf{Jaccard Index}} & \multicolumn{2}{c}{\textbf{Recall}} \\
& & HJI & NJI & OJI & SJI & HR & SR \\
\midrule
\scriptsize\parbox[t]{2mm}{\multirow{5}{*}{\rotatebox{90}{SAM}}} & English words & 4.2 & 12.1 & 48.9 & 11.1 & 5.4 & 30.4 \\
& WordNet & 4.9 & 14.5 & 48.2 & 12.5 & 6.1 & 35.9 \\
& BLIP-2 (ViT-L) & 15.8 & 19.2 & 31.7 & 17.2 & 33.4 & 60.6 \\
& BLIP-2 (ViT-g) & 15.1 & 17.5 & 27.2 & 16.4 & 33.1 & 61.7 \\
& LLaVA 1.5 (7B) & 18.3 & 18.9 & 29.3 & 17.5 & \textbf{41.6} & \textbf{63.6} \\
\midrule
\scriptsize\parbox[t]{2mm}{\multirow{2}{*}{\rotatebox{90}{SAN}}} & English words & 5.8 & 18.6 & 48.4 & 13.8 & 7.2 & 38.1 \\
& WordNet & 7.2 & 19.3 & 46.4 & 15.7 & 8.7 & 45.3 \\
\midrule
\rowcolor{ForestGreen!20} \multicolumn{2}{c|}{SAM + \methodShort} & 13.7 & 17.5 & 44.6 & 15.5 & 20.1 & 48.0 \\
\rowcolor{ForestGreen!20} \multicolumn{2}{c|}{SAN + \methodShort} & \textbf{26.9} & \textbf{30.2} & \textbf{53.2} & \textbf{20.8} & 34.2 & 61.8 \\
\rowcolor{ForestGreen!20} \multicolumn{2}{c|}{\methodSShort} & 20.5 & 20.3 & 32.2 & 18.5 & 31.4 & 61.8 \\
\bottomrule
\end{tabular}
\vspace{1mm}
\caption{Semantic segmentation on PascalVOC-20. \inlineColorbox{ForestGreen!20}{Green} are our methods.}
\label{tab:metric_segmentation_voc}
\end{table}

\begin{table}[t]
\centering
\begin{tabular}{cc|cccc|cc}
\toprule
\multicolumn{2}{c|}{\textbf{Method}}  & \multicolumn{6}{c}{\textbf{CTX-59}} \\
& & \multicolumn{4}{c|}{\textbf{Jaccard Index}} & \multicolumn{2}{c}{\textbf{Recall}} \\
& & HJI & NJI & OJI & SJI & HR & SR \\
\midrule
\scriptsize\parbox[t]{2mm}{\multirow{5}{*}{\rotatebox{90}{SAM}}} & English words & 1.0 & 6.4 & \textbf{40.2} & 8.9 & 2.5 & 27.0 \\
& WordNet & 1.0 & 8.7 & 39.6 & 9.6 & 2.5 & 28.9 \\
& BLIP-2 (ViT-L) & 7.0 & 10.5 & 28.4 & 12.1 & 16.1 & 42.7 \\
& BLIP-2 (ViT-g) & 6.5 & 10.3 & 27.4 & 12.0 & 15.5 & 44.4 \\
& LLaVA 1.5 (7B) & 9.1 & 12.0 & 29.7 & 12.9 & \textbf{21.4} & 48.3 \\
\midrule
\scriptsize\parbox[t]{2mm}{\multirow{2}{*}{\rotatebox{90}{SAN}}} & English words & 1.2 & 8.0 & 34.9 & 10.3 & 2.7 & 30.7 \\
& WordNet & 2.0 & 9.4 & 33.0 & 11.2 & 3.6 & 33.6 \\
\midrule
\rowcolor{ForestGreen!20} \multicolumn{2}{c|}{SAM + \methodShort} & 7.5 & 11.0 & 38.2 & 11.5 & 11.2 & 39.4 \\
\rowcolor{ForestGreen!20} \multicolumn{2}{c|}{SAN + \methodShort} & \textbf{15.5} & \textbf{16.2} & 38.1 & \textbf{14.7} & 20.8 & 46.9 \\
\rowcolor{ForestGreen!20} \multicolumn{2}{c|}{\methodSShort} & 13.4 & 13.1 & 32.6 & 13.9 & 20.8 & \textbf{48.6} \\
\bottomrule
\end{tabular}
\vspace{1mm}
\caption{Segmentation on PASCAL Context-59. \inlineColorbox{ForestGreen!20}{Green} highlights our methods.}
\label{tab:metric_segmentation_ctx}
\end{table}

\begin{table}[t]
\centering
\begin{tabular}{cc|cccc|cc}
\toprule
\multicolumn{2}{c|}{\textbf{Method}}  & \multicolumn{6}{c}{\textbf{ADE-150}} \\
& & \multicolumn{4}{c|}{\textbf{Jaccard Index}} & \multicolumn{2}{c}{\textbf{Recall}} \\
& & HJI & NJI & OJI & SJI & HR & SR \\
\midrule
\scriptsize\parbox[t]{2mm}{\multirow{5}{*}{\rotatebox{90}{SAM}}} & English words & 2.2 & 4.5 & \textbf{31.2} & 6.3 & 3.3 & 27.4 \\
& WordNet & 2.3 & 6.3 & 30.9 & 6.4 & 3.5 & 27.45 \\
& BLIP-2 (ViT-L) & 5.0 & 7.7 & 24.7 & 7.7 & 10.5 & 36.5 \\
& BLIP-2 (ViT-g) & 4.9 & 8.2 & 23.9 & 7.7 & 10.9 & 37.8 \\
& LLaVA 1.5 (7B) & 6.2 & 7.1 & 24.2 & 7.8 & 14.4 & 40.4 \\
\midrule
\scriptsize\parbox[t]{2mm}{\multirow{2}{*}{\rotatebox{90}{SAN}}} & English words & 2.2 & 4.5 & 19.2 & 7.2 & 3.6 & 30.5 \\
& WordNet & 2.8 & 5.1 & 19.1 & 7.5 & 4.8 & 32.9 \\
\midrule
\rowcolor{ForestGreen!20} \multicolumn{2}{c|}{SAM + \methodShort} & 6.1 & 7.7 & 29.4 & 7.4 & 10.0 & 35.2 \\
\rowcolor{ForestGreen!20} \multicolumn{2}{c|}{SAN + \methodShort} & 7.2 & 7.5 & 20.8 & 8.7 & 11.3 & 40.9 \\
\rowcolor{ForestGreen!20} \multicolumn{2}{c|}{\methodSShort} & \textbf{8.6} & \textbf{9.1} & 24.1 & \textbf{8.8} & \textbf{16.8} & \textbf{43.6} \\
\bottomrule
\end{tabular}
\vspace{1mm}
\caption{Semantic segmentation on ADE20K-150. \inlineColorbox{ForestGreen!20}{Green} are our methods.}
\label{tab:metric_segmentation_ade}
\end{table}

\begin{table*}
\centering
\begin{tabular}[b]{cc|ccc|ccc|c|c}
\toprule
\textbf{Database}  & \textbf{Size} & \textbf{CA}& \textbf{S-Sim.} &\textbf{ S-IoU} & \textbf{Nouns} (\%) & \textbf{Adjs} (\%) & \textbf{Verbs} (\%) & \textbf{Concepts} &  C.C. S-Sim. \\\midrule
  Ade20K & 0.07M & 11.0 & 31.8 & 1.7 & 82.7 & 10.6 & 6.7 & 3.1K & 19.0 \\
  COCO & 0.9M & 13.8 & 35.6 & 2.6 & 86.1 & 9.5 & 4.4 & 14.6K & 22.6 \\
  SBU Captions & 1.0M & 20.1 & 40.8 & 5.7 & 98.0 & 1.0 & 0.9 & 29.3K & 26.9 \\
  OpenImages & 1.4M & 16.8 & 37.8 & 4.2 & 85.4 & 10.0 & 4.5 & 21.4K & 24.8 \\
  Loc. Narr. & 1.9M & 15.7 & 37.1 & 4.1 & 86.9 & 9.8 & 3.2 & 30.9K & 24.9 \\
  \rowcolor{CornflowerBlue!20} CC3M & 2.8M & 37.9 & 53.9 & 16.2 & 57.9 & 23.6 & 18.5 & 28.4K & 35.2\\
  \rowcolor{CornflowerBlue!20} WIT & 4.8M & 32.9 & 47.9 & 14.5 & 99.8 & 0.05 & 0.07 & 163.7K & 30.0 \\
  Visual Genome & 5.4M & 14.6 & 35.2 & 3.8 & 75.1 & 15.6 & 9.2 & 21.1K & 26.7 \\
  \rowcolor{CornflowerBlue!20} Redcaps & 7.9M & 41.1 & 54.7 & 19.6 & 98.6 & 0.7 & 0.6 & 50.5K & 37.5 \\
  \rowcolor{CornflowerBlue!20} CC12M & 10.3M  & \textbf{43.8}& \textbf{57.4}& \textbf{21.2} & 96.6 & 1.8 & 1.5 & 36.7K & \textbf{38.7} \\
  \rowcolor{CornflowerBlue!20} YFCC100M* & 29.9M & 39.1 & 53.5 & 18.9 & 99.7 & 0.2 & 0.1 & 179.6K & \underline{37.8} \\
  \rowcolor{ForestGreen!20} Ours & 54.8M & \underline{41.5} & \underline{55.7} & \underline{20.4} & 99.8 & 0.05 & 0.05 & 334.1K & \textbf{38.7} \\
  LAION-400M & 413.8M & 37.7 & 52.7 & 18.7 & 99.4 & 0.5 & 0.1 & 1.68M & 37.3 \\
\bottomrule
\end{tabular}
\vspace{1pt}
\caption{ImageNet accuracy of \methodShort\ for different databases. \inlineColorbox{CornflowerBlue!20}{Blue} are the top-5 ones of PMD, used for \inlineColorbox{ForestGreen!20}{Ours}. \textit{C.C.~S-Sim.} is the semantic similarity of the closest caption to each image in the dataset. \textbf{Bold} is best, while \underline{underline} second best. YFCC100M* denotes the same subset used in PMD.}
\label{tab:ablation_knowledge_base}
\end{table*}

\begin{table}[t]
\centering
\begin{tabular}{cc|ccc}
\toprule
\textbf{Vocabulary} & \textbf{Prompt ensemble} & \textbf{CA} & \textbf{S-Sim.} & \textbf{S-IoU} \\
\midrule
\multirow{2}{*}{WordNet} &  & 32.6 & 40.2 & 9.1 \\
 & \ding{51} & 33.3 & 41.8 & 10.1 \\
\multirow{2}{*}{English Words} &  & 30.4 & 32.5 & 5.9 \\
 & \ding{51} & 29.6 & 34.3 & 7.0 \\
\rowcolor{ForestGreen!20} & & 43.1 & 50.4 & 17.7 \\
\rowcolor{ForestGreen!20}\multirow{-2}{*}{Ours} & \ding{51} & 44.1 & 50.6 & 18.3 \\
\bottomrule
\end{tabular}
\vspace{1mm}
\caption{Ablation on prompt ensembling. Results are averaged on the ten classification datasets. \inlineColorbox{ForestGreen!20}{Green} is our configuration.}
\label{tab:ablation_multiprompt}
\end{table}

\begin{table}[t]
\centering
\begin{tabular}{cc|ccc}
\toprule
\textbf{Backbone} & \textbf{Size (M)} & \textbf{CA} & \textbf{S-Sim.} & \textbf{S-IoU} \\
\midrule
ResNet-50 & 102 & 32.7 & 50.1 & 15.4 \\
\rowcolor{ForestGreen!20} ViT-L/14 & 427 & 43.5 & 56.9 & 21.2 \\
ViT-L/16 (SigLIP-384) & 652 & \textbf{47.8} & \textbf{59.4} & \textbf{22.5} \\
\bottomrule
\end{tabular}
\vspace{1mm}
\caption{Ablation on \methodShort\ backbone. Results are collected on ImageNet, using CC12M as the text database. Sizes are taken from \cite{ilharco_gabriel_2021_5143773}.}
\label{tab:ablation_scaling_cased}
\end{table}

\noindent\textbf{Quantitative results.}
We report results on Tab.~\ref{tab:metric_segmentation_voc}, Tab.~\ref{tab:metric_segmentation_ctx}, and Tab.~\ref{tab:metric_segmentation_ade}. Our experiments consistently demonstrate that English Words and WordNet are sub-optimal for the task, independently of whether a method first segment the image into regions of interest and then classify (\ie, SAM-based methods), or first generates candidate names and then segment (\ie, SAN-based methods). With SAM, the best approach exploits LLaVA 1.5 (7B) followed by BLIP-2 (both ViT-L and ViT-g), generally achieving comparable results across the six metrics and three datasets, with, \eg, $12.7$ and $12.3$ on NJI, and $50.8$ and $47.3$ on SR, respectively. The strongest method consists of using SAN with \methodShort\ to generate a custom list of candidates for each image and then segment with the open-vocabulary segmentation model. When compared with more naive approaches, \ie, English Words or WordNet for retrieval, SAN with \methodShort\ improves by, \eg, about $+12.9$ on HJI and $+17.0$ on HR. SAN with \methodShort\ and the best SAM-based approaches generally performs comparably on recall metrics, whether hard or soft. Notably, \methodSShort, despite not including any component pre-trained for semantic segmentation, achieves and sometimes surpasses all the SAM-based approaches, even achieving the highest scores on hard and soft recall, \ie, $+0.9\%$ and $+1.4\%$ against SAN with \methodShort. Due to the coarse nature of the segmentation mask, however, \methodSShort\ falls behind against methods trained for open-vocabulary semantic segmentation, while keeping an edge against SAM-based methods. Specifically, it achieves $-2.3$ and $-1.0$ on HJI and SJI against SAN with \methodShort, but improves on average by $+4.4$ and $+1.8$ against SAM with LLaVA (7B) or BLIP-2.

\subsection{Ablation studies}

In this section, we study various aspects of our approach. First, we validate the impact of the retrieval database. Then, we demonstrate how prompt ensemble improves the performance of retrieval-based baselines compared to image-text comparison. Lastly, we analyze how the backbone and training objective impact the performance of \methodShort.

\noindent\textbf{Retrieval database.}
We examine the effects of sourcing captions from different databases with various noise levels and sizes (\eg, from 0.07M captions of Ade20K to 413.8M of LAION-400M). For each database, we report cluster accuracy, semantic similarity and semantic IoU on ImageNet, alongside dataset statistics, \ie, percentage of nouns, adjectives, verbs, and number of concepts. Last, we also report the semantic similarity of the closest caption (C.C. S-Sim.) for each ImageNet sample. As indicated in Tab.~\ref{tab:ablation_knowledge_base}, there is a general trend of improved results with increasing database size. However, the quality of the captions plays also role a role, as seen from CC12M and Redcaps, achieving results that are either on par with or slightly superior to our database, combining CC3M, WIT, Redcaps, CC12M, and a subset of YFCC100M.
Interestingly, the average performance generally correlates with the C.C. S-Sim., hinting is as a possible criterion for database selection in \taskShort\ and \taskSShort.

\noindent\textbf{Prompt ensemble.}
We report the performance of prompt ensembling on all the retrieval-based baselines, \ie, English Words, WordNet, and \methodShort in Tab.~\ref{tab:ablation_multiprompt}. The results show consistent improvements across all baselines when using multiple templates. This outcome is consistent across the three metrics, except for one configuration, \ie, cluster accuracy for English Words. On average, prompt ensemble improves performance by $+0.3\%$ on cluster accuracy, $+1.2$ on semantic similarity, and $+0.9$ on semantic IoU.

\noindent\textbf{Scaling \methodShort\ backbone.} 
We test \methodShort\ considering different backbones. Specifically, we employ ResNet-50, ViT-L/14, and ViT-L/16. The first two backbones are pre-trained on 400M image-text samples by OpenAI, while the latter is trained with a sigmoid loss~\cite{zhai2023sigmoid} on WebLI~\cite{chen2022pali}. For this analyses, we use the smaller CC12M as retrieval database instead of PMD, as they achieve comparable performance in Tab.~\ref{tab:ablation_knowledge_base}. We report the results in Tab.~\ref{tab:ablation_scaling_cased} Notably, our approach shows positive changes when equipped with an higher number of parameters and better training losses, hinting that larger VLMs can lead to higher \taskShort\ performance.

\section{Conclusions}
\label{sec:discussion}

In this work, we formalise two tasks, \taskName\ and \taskSName, both operating on an unconstrained semantic space, tackling classification at image- and pixel-level without a pre-defined set of classes. We propose a method to tackle the first, \methodShort, confirming its efficacy across a wide range of benchmark datasets. In addition, we present an upgraded version of \methodShort\ for classification and a modification enabling dense image classification. We term the two variations \methodEShort\ and \methodSShort. We confirm the validity of our approaches across a wide range of benchmark datasets and evaluation metrics. For both the tasks, we present a suite of metrics designed for the openness of classification and semantic segmentation for vocabulary-free settings.

\noindent\textbf{Limitations and future works.} The efficacy of the \methodShort\ family of methods is strongly influenced by the selection of the retrieval database, with potential challenges in retrieving concepts that are not adequately represented therein. For instance, if the application domain encompasses fine-grained concepts, a generic database may not be appropriate. Similarly, \methodShort\ may mirror potential biases in the dataset. Nevertheless, we can flexibly address these issues by progressively incorporating new concepts into the database (including domain-specific ones) from textual corpora, without the need for retraining. Enhancements in bias mitigation and data quality control could alleviate this problem. Future studies could investigate strategies for automatically selecting or expanding a database based on test samples.

Both \methodShort\ and \methodSShort\ (or variants) do not maintain a record of their output history, which could lead to inconsistent predictions. This can lead in \taskShort\ to assign slightly different labels to images of the same semantic concept (for example, \textit{cassowary} versus \textit{Casuarius}). In \taskSShort\ this may even lead to inconsistencies at the image-level, with distant pixels of the same category receiving different labels (\eg, \textit{ground} vs \textit{floor}). Implementing a memory feature in \methodShort\ that stores the predicted labels could resolve this issue for classification, while for segmentation predictions could benefit from post-processing techniques merging regions with semantically and visually similar features. Lastly, \methodShort\ and its variants do not handle class granularity. For instance, in classification a \textit{cassowary} could also be predicted as a \textit{bird}. In segmentation, this may require understanding the level of detail of the masks (\eg, parts vs whole). Future research could clarify such instances by explicitly incorporating user needs into the \taskShort\ and \taskSShort\ family of models. Finally, while our \methodSShort\ does not use a segmentation network and does not require training, it performs multiple forwards of the same input image to generate a dense feature map. This is not computationally efficient at inference time and provides only coarse spatial information. The next steps for \methodSShort\ could develop tailored, even training-based, paradigms to produce finer dense feature maps faster.

{\small\bibliographystyle{IEEEtran}\bibliography{main}}
\begin{IEEEbiography}[{\includegraphics[width=0.9in,clip,keepaspectratio]{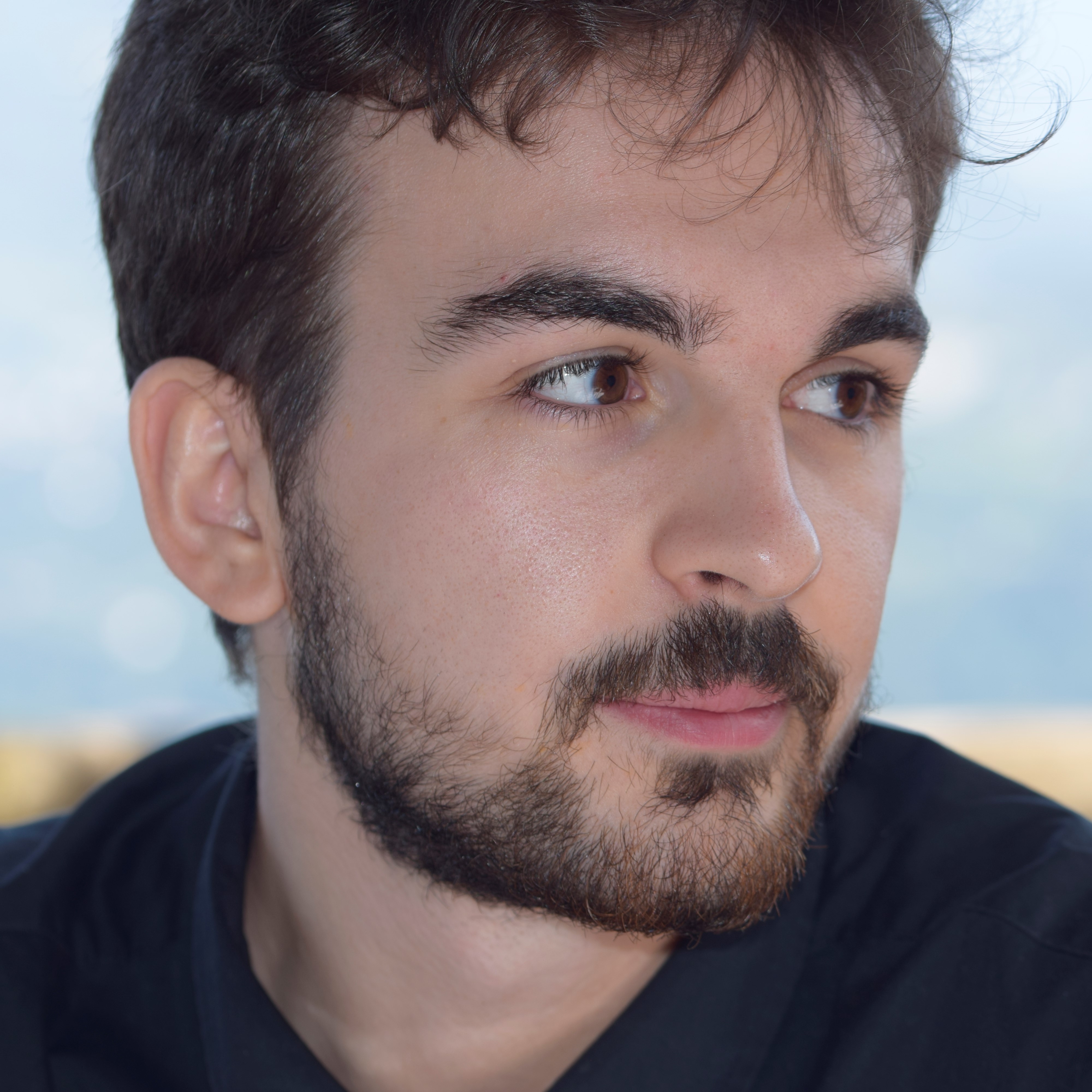}}]{Alessandro Conti}
is a PhD student at the University of Trento, in the Multimedia and Human Understanding Group (MHUG).
His research interests include multimodal learning, self- and weakly-supervised learning, and domain adaptation.
\end{IEEEbiography}

\begin{IEEEbiography}[{\includegraphics[width=0.9in,clip,keepaspectratio]{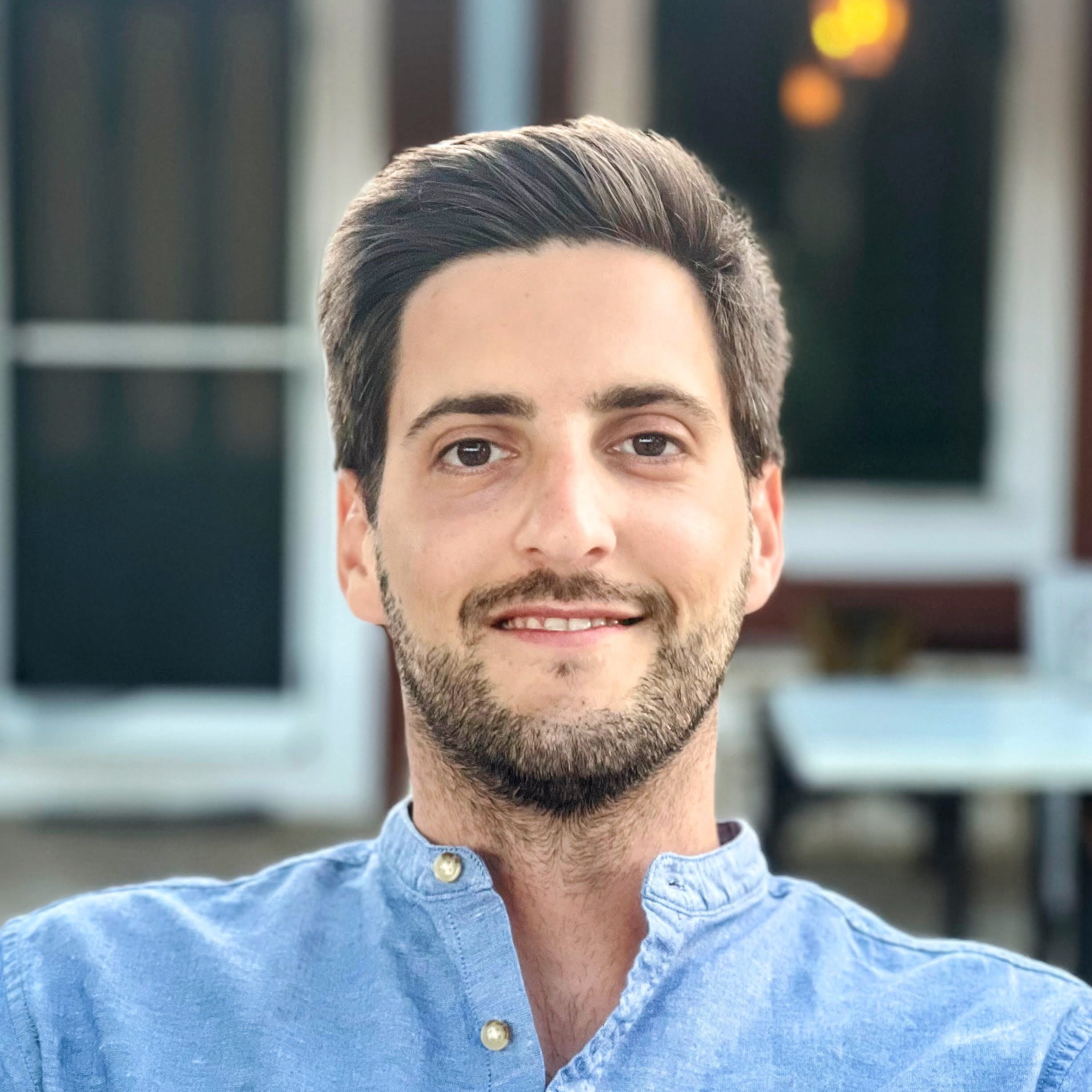}}]{Enrico Fini}
is a Research Scientist at Apple in Zurich.
He completed his Ph.D. at the University of Trento in the Multimedia and Human Understanding Group (MHUG).
During his Ph.D., Enrico interned at FAIR (Meta AI), Amazon, SAP AI Research, and Inria Grenoble.
His research interests include self-supervised learning, vision-language pre-training, continual learning, and open-set recognition.
\end{IEEEbiography}

\begin{IEEEbiography}[{\includegraphics[width=0.9in,clip,keepaspectratio]{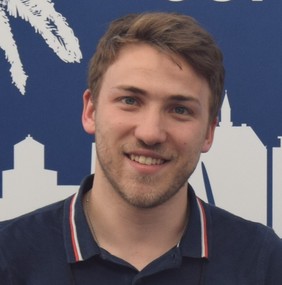}}]{Massimiliano Mancini}
is an assistant professor at the University of Trento, in the Multimedia and Human Understanding Group.
He completed his Ph.D. at the Sapienza University of Rome, and was a postdoc at the Cluster of Excellence ML, University of Tübingen.
He was a member of the ELLIS Ph.D. program, the TeV lab at Fondazione Bruno Kessler, and the VANDAL lab of the Italian Institute of Technology.
His research interests include transfer learning and compositionality.
\end{IEEEbiography}

\begin{IEEEbiography}[{\includegraphics[width=0.9in,clip,keepaspectratio]{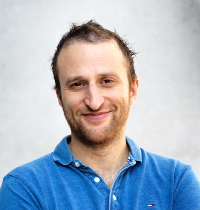}}]{Paolo Rota}
is an assistant professor at the Center for Mind and Brain (CIMeC) in the University of Trento.
He received his Ph.D. from the same university and has worked as a postdoctoral Marie Curie fellow at TU Wien and as a postdoc at the Istituto Italiano di Tecnologia in Genoa.
He also worked as an ML researcher at the ProM Facility in Rovereto.
He has been working as an assistant professor at the University of Trento since 2019 and started his tenure track in 2022.
His research interests are focused on image and video classification using Vision and Language.

\end{IEEEbiography}

\begin{IEEEbiography}[{\includegraphics[width=0.9in,clip,keepaspectratio]{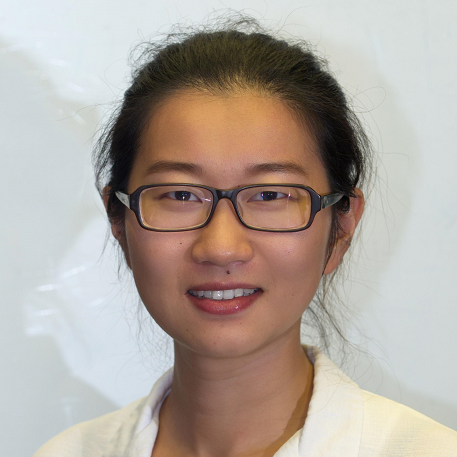}}]{Yiming Wang}
is a researcher in the Deep Visual Learning (DVL) unit at Fondazione Bruno Kessler (FBK).
Yiming obtained her PhD in Electric Engineering from Queen Mary University of London (UK) in 2018.
She is a member of ELLIS and an associate editor of International Journal of Social Robotics.
She has expertise in robotic perception and scene understanding.
Her recent research focuses on training-free methods addressing open-world recognition.
\end{IEEEbiography}

\begin{IEEEbiography}[{\includegraphics[width=0.9in,clip,keepaspectratio]{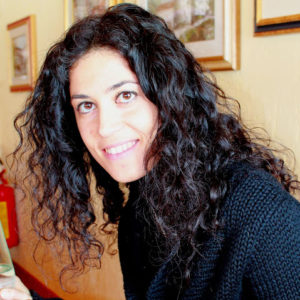}}]{Elisa Ricci}
is an associate professor with the University of Trento and a head of research unit with Fondazione Bruno Kessler.
She received the Honorable mention award at ICCV 2021 and the Best Paper Award at ACM MM 2021.
Her research interests are mainly in the areas of computer vision and deep learning.
She is an ELLIS fellow.
\end{IEEEbiography}

\newpage
\newpage

\appendix

\section*{A. Semantic space representation}
\label{sec:semantic_space_rep}

As the main challenge of \taskShort\ and \taskSShort, how to represent the large semantic space plays a fundamental role in the method design.
We can either model the multimodal semantic space directly with a VLM equipped with an autoregressive language decoder~\cite{li2022blip} or via image-text retrieval from VLDs.
Consequently, we can approach vocabulary-free tasks either via VQA-enabled VLMs by querying for the candidate class given the input image, or by retrieving and processing data from an external VLD to obtain the candidate class.

To investigate the two potential strategies, we perform a preliminary experimental analysis to understand how well the output of a method semantically captures the image category, or in other words, to assess the alignment of class boundaries in the visual and textual representations in \taskShort.
Specifically, we compare the semantic accuracy of querying VQA VLMs and of retrieving from VLDs w.r.t.~the ground-truth class labels.
We consider the output of a method as correct if its closest textual embedding among the target classes of the dataset corresponds to the ground-truth class of the test sample\footnote{Note that this metric is not the standard accuracy in image classification as we use distances in the embedding space to ground predictions from the unconstrained semantic space to the set of classes in a specific dataset.}.
We exploit the text encoder of CLIP (ViT-L)~\cite{clip} to obtain textual embeddings.

Regarding experimented methods, we select BLIP-2~\cite{li2023blip} to represent VQA-enabled VLMs for its state-of-the-art performance in VQA benchmarks, while we use a subset of PMD~\cite{pmd} as the VLD.
In particular, we compare the following methods: i) \textit{BLIP-2 VQA}, which directly queries BLIP-2 for the image category; ii) \textit{BLIP-2 Captioning}, which queries BLIP-2 for the image caption; iii) \textit{Closest Caption}, which is the closest caption to the image, as retrieved from the database; iv) \textit{Caption Centroid}, which averages the textual embeddings of the 10 most similar captions to the input image.
As we use CLIP embeddings, if visual and textual representations perfectly align, the performance would be the same as zero-shot CLIP with given target classes.
We thus report zero-shot CLIP to serve as the upper bound for retrieval accuracy.

\begin{figure*}[t!]
  \centering
  \includegraphics[width=1.0\textwidth]{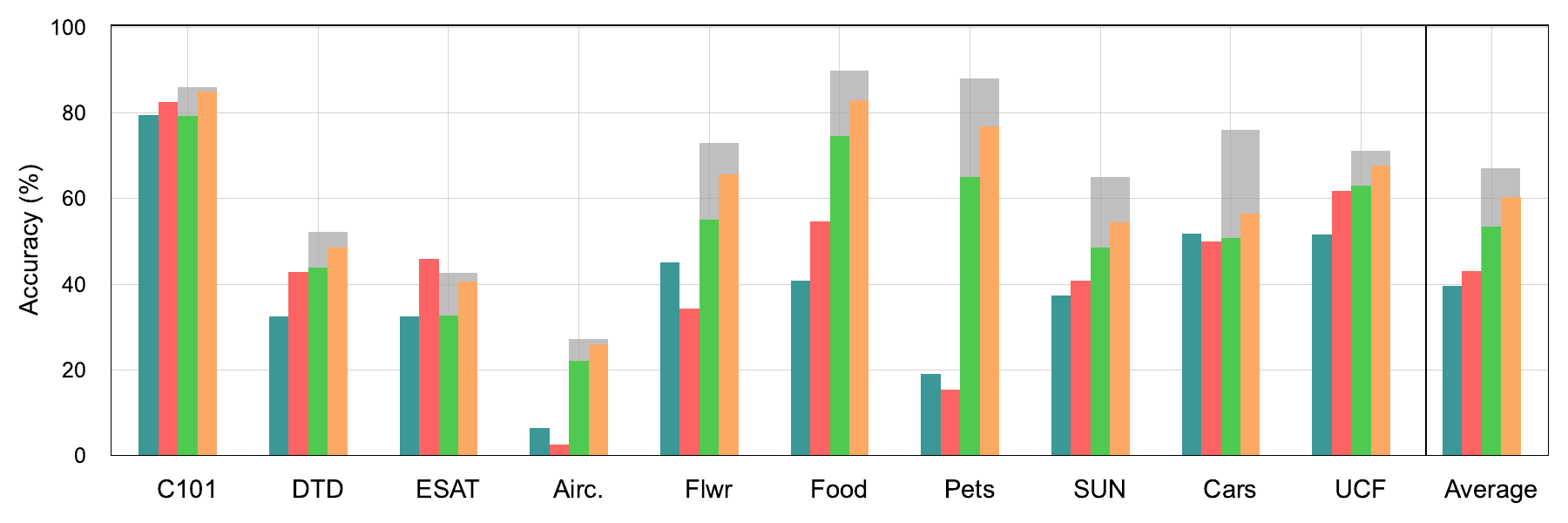}
  \caption{Results of our preliminary study, showing the top-1 accuracy when matching semantic descriptions to ground-truth class names in ten different datasets.
  We compare \inlineColorbox{BlueEnrico!40}{BLIP-2~(VQA)} and \inlineColorbox{RedEnrico!40}{BLIP-2~(Captioning)} with \inlineColorbox{GreenEnrico!40}{Closest~Caption} and \inlineColorbox{OrangeEnrico!40}{Captions~Centroid}, \ie, the average representation of the retrieved captions.
  We additionally highlight the \inlineColorbox{Gray!30}{Upper~bound} for zero-shot CLIP.
  Representing the large semantic space as VLDs and retrieving captions from it produces semantically more similar outputs to ground-truth labels w.r.t.~querying outputs from VQA-enabled VLMs, while requiring 10 times fewer parameters compared to the latter.
  }
  \label{fig:pilot_study}
\end{figure*}

We experiment on a variety of test datasets for both coarse- and fine-grained classification (see details in Sec.~\ref{sec:experiments}), and report the results in Fig.~\ref{fig:pilot_study}.
The average textual embedding of the retrieved captions (\ie~Caption Centroid) achieves the best semantic accuracy for 9 datasets out of 10, consistently surpassing methods based on BLIP-2.
On average, the accuracy achieved by Caption Centroid is $60.47\%$, which is $+17.36\%$ higher than the one achieved by BLIP-2 Captioning ($43.11\%$).
Moreover, Captions Centroid achieves results much closer to the CLIP upper bound ($67.17\%$) than the other approaches.
Notably, such VLD-based retrieval is also computationally more efficient, faster (approximately 2 seconds for a batch size of 64 on a single A6000 GPU), and requires fewer parameters (approximately 10 times less) than BLIP-2 (see Tab.~\ref{tab:cost} in Appendix \ref{sec:cost}).

The results of this preliminary study clearly suggest that representing the large semantic space with VLDs can produce (via retrieval) semantically more relevant content to the input image, in comparison to querying VQA-enabled VLMs, while being computationally efficient.
Based on this conclusion, we develop an approach, \methodName~(\methodShort), that searches for the semantic class from the large semantic space represented in the captions of VLDs.

\section*{B. Details on BLIP-2 and LLaVA prompting}
\label{sec:prompting-blip}
In the main paper, we compare \methodShort\ and its variants by replacing the candidate generation module and prediction directly with a VLM with VQA capabilities, testing BLIP-2~\cite{li2023blip} and LLaVA 1.5 (7B)~\cite{liu2023improved}. Here we provide details on how we prompt these models.

{In line with the BLIP-2~\cite{li2023blip} demo, for captioning, we used the prompt "Question: what's in the image? Answer:". For VQA, we used "Question: what's the name of the object in the image? Answer: a".}
For LLaVA 1.5, we use the same prompts of BLIP-2, but append an "Omit any superfluous text." to avoid introductory or closing statements.

\section*{C. Candidates filtering}
\label{sec:candidates}

With the closest captions retrieved from the external database given the input image (please refer to Sec.~\ref{sec:ours} of the main manuscript for further details), we post-process them to filter out a set of candidate category names.
We first create a set of all words that are contained in the captions.
Then we apply sequentially three different groups of operations to the set to (i) remove noisy candidates, (ii) standardize their format, and (iii) filter them.

With the first group of operations, we remove all the irrelevant textual contents, such as tokens (\ie, "\textlangle~PERSON~\textrangle"), URLs, or file extensions.
Note that, for the file extensions, we remove the extension but retain the file name as it might contain candidate class names.
We also remove all the words that are shorter than three characters and split compound words by underscores or dashes.
Finally, we remove all those terms containing symbols or numbers and meta words that are irrelevant to the classification task, such as "image", "photo", or "thumbnail".
As shown in Table~\ref{tab:ablation_filtering}, when compared to having no operation for candidate filtering (first row), this set of operations removes inappropriate content and increases the accuracy of clusters by $+6.4\%$ and improves the semantic IoU by $+0.3$.
However, we can observe a drop in semantic similarity by $-1.5$.
This might be due to the removal of unnatural words that could still describe well the content of the image, \ie, underline- or dash-separated words, or URLs since they are longer w.r.t.~natural words.

The second group of operations standardize the candidate names by aligning words that refer to the same semantic class to a standard format, reducing class redundancy.
For example, "cassowary" and "Cassowary" will be considered as a single class instead of two.
To this end, we perform two operations—lowercase conversion and singular form conversion.
With such standardizing conversions, we observe a sizeable boost in terms of performances when compared to the results obtained by applying only removal-related operations.
As shown in Table~\ref{tab:ablation_filtering}, we achieve higher results across all three metrics, leading to a relative improvement of $+7.2\%$, $+0.7$, and $+1.2$ in terms of cluster accuracy, semantic similarity, and semantic IoU, respectively.

The last group of operations considers two forms of filtering, where the first aims to filter out entire categories of words via Part-Of-Speech (POS) tag and the second aims to filter out rare and noisy contents based on the word occurrences.
We select these two operations since common dataset class names do not contain terms that carry no semantics, \eg, articles and pronouns, and since~\cite{cui2022democratizing} showed that CLIP performs better when exposed to a smaller amount of unique tokens.
The POS tagging\footnote{We use the NLP library flair (\url{https://github.com/flairNLP/flair}).} categorizes words into groups, such as adjectives, articles, nouns, or verbs, enabling us to filter all the terms that are not semantically meaningful for a classification task.
Regarding the occurrence filtering, we first count how often a word appears in the retrieved captions and then we remove words that appear only once to make the candidate list less noisy.
We can see from Table~\ref{tab:ablation_filtering} that the inclusion of this final set of operations scores the best among all three metrics when compared to the results obtained when only the previous two groups of operations are applied.

\noindent\textbf{Number of captions vs number of selected candidates.} 
To complement the previous analysis, in Tab.~\ref{tab:num_captions-candidates}, we report the number of unique candidates extracted by the candidate filtering procedure, averaged over the ten datasets, and with an increasing number of retrieved captions, \ie, 1, 2, 5, 10, and 20. In the table, we show both the number of candidates extracted and the number of selected words. As the number of retrieved captions increases, the unique number of candidate words also increases, \ie, from 3849 with 1 caption to 28781 with 20. The number of selected words, however, stabilizes around 800 as soon as we retrieve more than 1 caption. Having more captions reduces the noises in the selected words, something that might be present when relying on a single caption.

\section*{D. Computational cost}
\label{sec:cost}
We analyze the computational efficiency of \methodShort~versus BLIP-2 and LLaVA 1.5 performing VQA and captioning, and report their respective number of parameters and inference time in Tab.~\ref{tab:cost}.
Notably, the methods using external databases are consistently faster than BLIP-2 and LLaVA.
For instance, \methodShort\ achieves a speed-up of 2x with respect to both the largest captioning model and the largest VQA model, while also achieving better performance.
In a similar fashion, our method is approximately 3.0x faster than LLaVA 1.5 for VQA and 3.5x for captioning.
Overall, the fastest method is Closest Caption, which exploits the external database to retrieve a single caption and does not consider any candidate extraction pipeline.
Conversely, our method retrieves the ten most similar captions and post-processes them to extract class name, resulting in a increase in inference time of approximately 2 times.
Compared with the CLIP upper bound, our method considers multiple additional steps, each adding extra inference time.
First, our method retrieves candidates from the external database to then extract the class names.
Second, we have to forward the class names through the text encoder for each sample, while CLIP can forward them once and cache the features for later reuse.

\begin{table}
\centering
\begin{tabular}[b]{cccccc}
\toprule
\multicolumn{3}{c}{\textbf{Operations}} & \multirow{2}[1]{*}{\textbf{CA}} & \multirow{2}[1]{*}{\textbf{S-Sim.}} & \multirow{2}[1]{*}{\textbf{S-IoU}} \\
\cmidrule(lr){1-3}
\textbf{Remove} & \textbf{Standardize} & \textbf{Filter} & & & \\ 
\midrule
& & & 27.7 &  48.8 & 15.0 \\
\ding{51} & & & 34.1 & 47.3 & 15.3 \\
\ding{51} & \ding{51} & & 41.3 & 48.0 & 16.5 \\
\ding{51} & \ding{51} & \ding{51} &  \textbf{43.1} & \textbf{50.4} & \textbf{17.7} \\
\bottomrule
\end{tabular}
\vspace{2pt}
\caption{Ablation on the candidate filtering operations. Metrics are averaged across the ten datasets.}
\label{tab:ablation_filtering}
\end{table}

\begin{table}[t]
\centering
\begin{tabular}{cc|ccc|cc}
\toprule
\multicolumn{2}{c|}{\textbf{Num. captions}} & \textbf{CA} & \textbf{S-Sim.} & \textbf{S-IoU} & \textbf{Candidates} & \textbf{Selected}\\
\midrule
& 1 & 30.9 & 43.2 & 12.4 & 3849  & 1047 \\
& 2 & 35.0 & 46.4 & 14.6 & 6867 & 854 \\
& 5 & 42.3 & \textbf{50.6} & 17.5 & 14548 & 775 \\
\rowcolor{ForestGreen!20} & 10 & \textbf{43.1} & 50.4 & \textbf{17.7} & 22475 & 794 \\
& 20 & 42.9 & 49.3 & 17.1  & 28781 & 802 \\
\bottomrule
\end{tabular}
\vspace{2pt}
\caption{Ablation on the number retrieved captions. \inlineColorbox{ForestGreen!20}{Green} represents our selected configuration. We expand the results of Fig.~4 in the main manuscript to show the number of unique words extracted from captions (\ie, ``candidates'') and the number of words selected by \methodShort~for classification (i.e., ``selected''). Results are averaged across the ten datasets.}
\label{tab:num_captions-candidates}
\end{table}

When using an external database, note that an increase in database size implies a minimal variation in retrieval time.
This is demonstrated by the computational cost required by retrieving from the large LAION-400M~\cite{laion} database.
As the results show, inference time is comparable between \methodShort\ and \methodShort~(LAION-400M) despite the latter being approximately 8 times larger.

\begin{table*}[t]
\small
\centering
\begin{tabular}{ccc|c}
\toprule
\multicolumn{2}{c}{\textbf{Method}} & \textbf{Num. Params.}  & \textbf{Inference time (ms) $\downarrow$} \\
\midrule
{\multirow{4}{*}{{Caption}}} & Closest Caption & 0.43B & 1390 ± 10 \\
 & BLIP-2 (ViT-L) & 3.46B & 5710 ± 153 \\
 & BLIP-2 (ViT-g) & 4.37B & 6870 ± 177 \\
 & LLaVA 1.5 (7B) & 7.49B & 8260 ± 20 \\
 \midrule
{\multirow{3}{*}{{VQA}}} & {BLIP-2 (ViT-L)} & 3.46B & 5670 ± 135 \\
 & {BLIP-2 (ViT-g)} & 4.37B & 6650 ± 117 \\
 & LLaVA 1.5 (7B) & 7.49B & 7040 ± 28 \\
\midrule
\rowcolor{ForestGreen!20} \multicolumn{2}{c}{\methodShort} & 0.43B & 2630 ± 14 \\
\rowcolor{ForestGreen!20} \multicolumn{2}{c}{\methodShort\ (LAION-400M)} & 0.43B & 2640 ± 11 \\
\midrule
\rowcolor{Gray!20}\multicolumn{2}{c}{CLIP upper bound} & 0.43B & 645 ± 77 \\
\bottomrule
\end{tabular}
\vspace{2pt}
\caption{Computational cost of different methods. \inlineColorbox{ForestGreen!20}{Green} is our method, \inlineColorbox{Gray!20}{gray} shows the upper bound. Inference time is reported on batches of size 64, as the average over multiple runs.}
\label{tab:cost}
\end{table*}

\begin{table*}[t]
\small
\begin{tabular}{ccc|cccccccccc|c}
\toprule
& \multicolumn{2}{c|}{\textbf{Method}}  &\multicolumn{11}{c}{\textbf{Cluster Accuracy (\%)} $\uparrow$} \\
& & & {C101} & {DTD} & {ESAT} & Airc. & Flwr & Food & {Pets} & {SUN} & {Cars} & {UCF} & \textbf{Avg.} \\
\midrule
\scriptsize\parbox[t]{2mm}{\multirow{7}{*}{\rotatebox{90}{CLIP RN50}}} & {\multirow{2}{*}{{CLIP}}} & WordNet & 30.3 & 18.3 & \textbf{22.5} & 13.2 & 47.8 & 31.4 & 45.2 & 26.0 & 14.2 & 31.2 & 28.0 \\
& & English Words & 24.8 & 17.5 & 18.5 & 13.4 & 49.5 & 23.1 & 36.6 & 22.2 & 15.5 & 27.1 & 24.8 \\
\cmidrule{2-14}
& {\multirow{1}{*}{{Caption}}} & Closest Caption & 9.7 & 7.1 & 13.3 & 8.4 & 21.2 & 6.2 & 8.7 & 6.5 & 12.8 & 14.9 & 10.9 \\
\cmidrule{2-14}
& \multicolumn{2}{c|}{\cellcolor{ForestGreen!20}\methodShort} & \textbf{44.6}\cellcolor{ForestGreen!20} & 23.9\cellcolor{ForestGreen!20} & 12.5\cellcolor{ForestGreen!20} & 15.3\cellcolor{ForestGreen!20} & 58.8\cellcolor{ForestGreen!20} & 48.7\cellcolor{ForestGreen!20} & 50.1\cellcolor{ForestGreen!20} & 32.8\cellcolor{ForestGreen!20} & 24.6\cellcolor{ForestGreen!20} & \textbf{33.9}\cellcolor{ForestGreen!20} & 34.5\cellcolor{ForestGreen!20} \\
& \multicolumn{2}{c|}{\cellcolor{ForestGreen!20}\methodEShort} & 44.4\cellcolor{ForestGreen!20} & \textbf{24.3}\cellcolor{ForestGreen!20} & 17.9\cellcolor{ForestGreen!20} & \textbf{\underline{16.7}}\cellcolor{ForestGreen!20} & \textbf{\underline{62.2}}\cellcolor{ForestGreen!20} &\textbf{\underline{49.1}}\cellcolor{ForestGreen!20} & \textbf{\underline{55.1}}\cellcolor{ForestGreen!20} & \textbf{33.5}\cellcolor{ForestGreen!20} & \textbf{29.0}\cellcolor{ForestGreen!20} & 33.8\cellcolor{ForestGreen!20} & \textbf{36.6}\cellcolor{ForestGreen!20} \\
\cmidrule{2-14}
& \multicolumn{2}{c|}{\cellcolor{Gray!20}CLIP upper bound} & 82.1\cellcolor{Gray!20} & 41.5\cellcolor{Gray!20} & 33.5\cellcolor{Gray!20} & 19.6\cellcolor{Gray!20} & 63.1\cellcolor{Gray!20} & 74.6\cellcolor{Gray!20} & 78.9\cellcolor{Gray!20} & 55.6\cellcolor{Gray!20} & 54.9\cellcolor{Gray!20} & 58.4\cellcolor{Gray!20} & 56.2\cellcolor{Gray!20} \\
\midrule
\scriptsize \parbox[t]{2mm}{\multirow{7}{*}{\rotatebox{90}{CLIP ViT-L/14}}} & {\multirow{2}{*}{{CLIP}}} & WordNet & 34.0 & 20.1 & 16.7 & 16.7 & 58.3 & 40.9 & 52.0 & 29.4 & 18.6 & 39.5 & 32.6\\
& & English Words & 29.1 & 19.6 & 22.1 & 15.9 & 64.0 & 30.9 & 44.4 & 24.2 & 19.3 & 34.5 & 30.4 \\
\cmidrule{2-14}
& {\multirow{1}{*}{{Caption}}} & Closest Caption & 12.8 & 8.9 & 16.7 & 13.3 & 28.5 & 13.1 & 15.0 & 8.6 & 20.0 & 17.8 & 15.5 \\
\cmidrule{2-14}
& \multicolumn{2}{c|}{\cellcolor{ForestGreen!20}\methodShort} & \textbf{51.5}\cellcolor{ForestGreen!20} & 29.1\cellcolor{ForestGreen!20} & \textbf{23.8}\cellcolor{ForestGreen!20} & 22.8\cellcolor{ForestGreen!20} & 68.7\cellcolor{ForestGreen!20} & 58.8\cellcolor{ForestGreen!20} & 60.4\cellcolor{ForestGreen!20} & 37.4\cellcolor{ForestGreen!20} & 31.3\cellcolor{ForestGreen!20} & \textbf{47.7}\cellcolor{ForestGreen!20} & 43.1\cellcolor{ForestGreen!20} \\
& \multicolumn{2}{c|}{\cellcolor{ForestGreen!20}\methodEShort} & 51.3\cellcolor{ForestGreen!20} & \textbf{29.3}\cellcolor{ForestGreen!20} & 21.0\cellcolor{ForestGreen!20} & \textbf{\underline{24.4}\cellcolor{ForestGreen!20}} & \textbf{\underline{70.4}}\cellcolor{ForestGreen!20} &\textbf{\underline{61.2}}\cellcolor{ForestGreen!20} & \textbf{\underline{60.9}}\cellcolor{ForestGreen!20} & \textbf{\underline{37.7}}\cellcolor{ForestGreen!20} & \textbf{38.5}\cellcolor{ForestGreen!20} & 46.6\cellcolor{ForestGreen!20} & \textbf{\underline{44.1}}\cellcolor{ForestGreen!20} \\
\cmidrule{2-14}
& \multicolumn{2}{c|}{\cellcolor{Gray!20}CLIP upper bound}& 87.6\cellcolor{Gray!20} & 52.9\cellcolor{Gray!20} & 47.4\cellcolor{Gray!20} & 31.8\cellcolor{Gray!20} & 78.0\cellcolor{Gray!20} & 89.9\cellcolor{Gray!20} & 88.0\cellcolor{Gray!20} & 65.3\cellcolor{Gray!20} & 76.5\cellcolor{Gray!20} & 72.5\cellcolor{Gray!20} & 69.0\cellcolor{Gray!20} \\
\midrule
\scriptsize\parbox[t]{2mm}{\multirow{7}{*}{\rotatebox{90}{SigLIP ViT-L/16}}} & {\multirow{2}{*}{{SigLIP}}} & WordNet & 35.0 & 23.0 & 26.5 & 20.7 & 59.2 & 52.4 & 59.8 & 30.8 & 37.7 & 39.1 & 38.4 \\
 &  & English Words & 28.6 & 21.4 & 21.4 & 19.0 & 60.3 & 41.2 & 53.6 & 24.1 & 38.0 & 33.8 & 34.1 \\
 \cmidrule{2-14}
 & {\multirow{1}{*}{{Caption}}} & Closest Caption & 15.6 & 9.7 & 26.2 & 23.9 & 36.5 & 19.3 & 17.2 & 8.9 & 36.3 & 26.9 & 22.0 \\
\cmidrule{2-14}
& \multicolumn{2}{c|}{\cellcolor{ForestGreen!20}\methodShort} & 56.3\cellcolor{ForestGreen!20} & \textbf{\underline{30.6}}\cellcolor{ForestGreen!20} & \textbf{32.3}\cellcolor{ForestGreen!20} & 24.6\cellcolor{ForestGreen!20} & 66.4\cellcolor{ForestGreen!20} & \textbf{\underline{67.0}}\cellcolor{ForestGreen!20} & \textbf{\underline{68.1}}\cellcolor{ForestGreen!20} & \textbf{\underline{40.2}}\cellcolor{ForestGreen!20} & \textbf{45.2}\cellcolor{ForestGreen!20} & 48.5\cellcolor{ForestGreen!20} & \textbf{\underline{47.9}}\cellcolor{ForestGreen!20} \\
& \multicolumn{2}{c|}{\cellcolor{ForestGreen!20}\methodEShort} & \textbf{58.1}\cellcolor{ForestGreen!20} & 29.5\cellcolor{ForestGreen!20} & 25.4\cellcolor{ForestGreen!20} & \textbf{\underline{29.5}}\cellcolor{ForestGreen!20} & \textbf{\underline{69.7}}\cellcolor{ForestGreen!20} & 62.8\cellcolor{ForestGreen!20} & 64.7\cellcolor{ForestGreen!20} & 39.9\cellcolor{ForestGreen!20} & 44.4\cellcolor{ForestGreen!20} & \textbf{49.6}\cellcolor{ForestGreen!20} & 47.4\cellcolor{ForestGreen!20} \\
\cmidrule{2-14}
& \multicolumn{2}{c|}{\cellcolor{Gray!20}CLIP upper bound} & 96.3\cellcolor{Gray!20} & 60.6\cellcolor{Gray!20} & 44.8\cellcolor{Gray!20} & 48.3\cellcolor{Gray!20} & 90.9\cellcolor{Gray!20} & 94.2\cellcolor{Gray!20} & 95.4\cellcolor{Gray!20} & 69.6\cellcolor{Gray!20} & 92.7\cellcolor{Gray!20} & 82.0\cellcolor{Gray!20} & 77.5\cellcolor{Gray!20} \\
\midrule
& {\multirow{3}{*}{{Caption}}} & BLIP-2 (ViT-L)& 26.5 & 11.7 & 23.3 & 5.4 & 23.6 & 12.4 & 11.6 & 19.5 & 14.8 & 25.7 & 17.4 \\
&  & BLIP-2 (ViT-g) & 37.4 & 13.0 & 25.2 & 10.0 & 29.5 & 19.9 & 15.5 & 21.5 & 27.9 & 32.7 & 23.3 \\
& & LLaVA 1.5 (7B) & 41.1 & 11.1 & 19.7 & 10.4 & 13.4 & 11.1 & 12.8 & 14.0 & 12.1 & 29.5 & 17.5 \\
\midrule
& {\multirow{3}{*}{{VQA}}} & {BLIP-2 (ViT-L)} & 60.4 & 20.4 & 21.4 & 8.1 & 36.7 & 21.3 & 14.0 & 32.6 & 28.8 & 44.3 & 28.8 \\
&  & {BLIP-2 (ViT-g)} & 62.2 & 23.8 & 22.0 & 15.9 & 57.8 & 33.4 & 23.4 & 36.4 & 57.2 & 55.4 & 38.7 \\
& & LLaVA 1.5 (7B) & 76.2 & 30.6 & 38.9 & 3.0 & 5.8 & 22.7 & 7.7 & 27.5 & 2.6 & 48.0 & 26.3 \\
\bottomrule
\end{tabular}
\vspace{2pt}
\caption{Cluster Accuracy on the ten datasets. \inlineColorbox{ForestGreen!20}{Green} is our method, \inlineColorbox{Gray!20}{gray} shows the upper bound. \textbf{Bold} represents best, \underline{underline} indicates best considering also image captioning and VQA models.}
\label{tab:extended_metric_semantic_cluster_acc}
\end{table*}

\begin{table*}[t]
\small
\begin{tabular}{ccc|cccccccccc|c}
\toprule
& \multicolumn{2}{c|}{\textbf{Method}}  &\multicolumn{11}{c}{\textbf{Semantic IoU (\%)} $\uparrow$} \\
& & & {C101} & {DTD} & {ESAT} & Airc. & Flwr & Food & {Pets} & {SUN} & {Cars} & {UCF} & \textbf{Avg.} \\
\midrule
\scriptsize \parbox[t]{2mm}{\multirow{6}{*}{\rotatebox{90}{CLIP RN50}}} & {\multirow{2}{*}{{CLIP}}} & WordNet & 9.8 & 1.9 & \textbf{0.9} & 0.0 & 21.2 & 5.5 & 14.1 & 7.3 & 3.5 & 2.6 & 6.7 \\
& & English Words & 5.1 & 0.8 & 0.0 & 0.1 & 12.4 & 1.8 & 13.2 & 7.8 & 2.5 & 1.3 & 4.5 \\
\cmidrule{2-14}
& {\multirow{1}{*}{{Caption}}} & Closest Caption & 3.4 & 0.5 & 0.1 & 1.5 & 6.6 & 2.2 & 2.2 & 2.1 & 9.1 & 0.7 & 2.8 \\
\cmidrule{2-14}
& \multicolumn{2}{c|}{\cellcolor{ForestGreen!20}\methodShort} & 31.1\cellcolor{ForestGreen!20} & 2.9\cellcolor{ForestGreen!20} & 0.1\cellcolor{ForestGreen!20} & \textbf{\underline{2.8}}\cellcolor{ForestGreen!20} & 29.7\cellcolor{ForestGreen!20} & 15.1\cellcolor{ForestGreen!20} & 27.6\cellcolor{ForestGreen!20} & 15.0\cellcolor{ForestGreen!20} & \textbf{13.4}\cellcolor{ForestGreen!20} & \textbf{\underline{5.2}}\cellcolor{ForestGreen!20} & 14.3\cellcolor{ForestGreen!20} \\
& \multicolumn{2}{c|}{\cellcolor{ForestGreen!20}\methodEShort} & \textbf{33.0}\cellcolor{ForestGreen!20} & \textbf{\underline{3.7}}\cellcolor{ForestGreen!20} & 0.4\cellcolor{ForestGreen!20} & 2.7\cellcolor{ForestGreen!20} & \textbf{30.4}\cellcolor{ForestGreen!20} & \textbf{\underline{15.2}}\cellcolor{ForestGreen!20} & \textbf{\underline{29.7}}\cellcolor{ForestGreen!20} & \textbf{\underline{15.7}}\cellcolor{ForestGreen!20} & 12.9\cellcolor{ForestGreen!20} & 4.7\cellcolor{ForestGreen!20} & \textbf{14.8}\cellcolor{ForestGreen!20} \\
\cmidrule{2-14}
& \multicolumn{2}{c|}{\cellcolor{Gray!20}CLIP upper bound} & 81.5\cellcolor{Gray!20} & 41.4\cellcolor{Gray!20} & 33.9\cellcolor{Gray!20} & 16.5\cellcolor{Gray!20} & 60.2\cellcolor{Gray!20} & 74.6\cellcolor{Gray!20} & 78.9\cellcolor{Gray!20} & 56.9\cellcolor{Gray!20} & 66.5\cellcolor{Gray!20} & 57.1\cellcolor{Gray!20} & 56.8\cellcolor{Gray!20} \\
\midrule
\scriptsize \parbox[t]{2mm}{\multirow{6}{*}{\rotatebox{90}{CLIP ViT-L/14}}} & {\multirow{2}{*}{{CLIP}}} & WordNet & 15.0 & 3.0 & 1.3 & 0.5 & 31.3 & 7.8 & 14.7 & 9.0 & 4.8 & 3.8 & 9.1 \\
& & English Words & 8.0 & 2.0 & 0.0 & 1.1 & 16.4 & 2.0 & 17.2 & 8.1 & 2.7 & 1.8 & 5.9 \\
\cmidrule{2-14}
& {\multirow{1}{*}{{Caption}}} & Closest Caption & 4.5 & 0.8 & 1.3 & 1.9 & 5.9 & 3.1 & 3.0 & 2.3 & 11.4 & 1.0 & 3.5 \\
\cmidrule{2-14}
& \multicolumn{2}{c|}{\cellcolor{ForestGreen!20}\methodShort} & 35.4\cellcolor{ForestGreen!20} & \textbf{\underline{5.1}}\cellcolor{ForestGreen!20} & 2.3\cellcolor{ForestGreen!20} & \textbf{\underline{4.8}}\cellcolor{ForestGreen!20} & 33.1\cellcolor{ForestGreen!20} & \textbf{\underline{19.4}}\cellcolor{ForestGreen!20} & \textbf{\underline{35.1}}\cellcolor{ForestGreen!20} & 17.2\cellcolor{ForestGreen!20} & \textbf{16.2}\cellcolor{ForestGreen!20} & \textbf{\underline{8.4}}\cellcolor{ForestGreen!20} & 17.7\cellcolor{ForestGreen!20} \\
& \multicolumn{2}{c|}{\cellcolor{ForestGreen!20}\methodEShort} & \textbf{37.8}\cellcolor{ForestGreen!20} & 4.6\cellcolor{ForestGreen!20} & \textbf{5.2}\cellcolor{ForestGreen!20} & 4.4\cellcolor{ForestGreen!20} & \textbf{35.2}\cellcolor{ForestGreen!20} & 18.9\cellcolor{ForestGreen!20} & 34.9\cellcolor{ForestGreen!20} & \textbf{\underline{17.8}}\cellcolor{ForestGreen!20} & 16.0\cellcolor{ForestGreen!20} & 7.8\cellcolor{ForestGreen!20} & \textbf{\underline{18.3}}\cellcolor{ForestGreen!20} \\
\cmidrule{2-14}
& \multicolumn{2}{c|}{\cellcolor{Gray!20}CLIP upper bound} & 86.0\cellcolor{Gray!20} & 52.2\cellcolor{Gray!20} & 51.5\cellcolor{Gray!20} & 28.6\cellcolor{Gray!20} & 75.7\cellcolor{Gray!20} & 89.9\cellcolor{Gray!20} & 88.0\cellcolor{Gray!20} & 66.6\cellcolor{Gray!20} & 84.5\cellcolor{Gray!20} & 71.3\cellcolor{Gray!20} & 69.4\cellcolor{Gray!20} \\
\midrule
\scriptsize\parbox[t]{2mm}{\multirow{7}{*}{\rotatebox{90}{SigLIP ViT-L/16}}} & {\multirow{2}{*}{{SigLIP}}} & WordNet & 18.0 & \textbf{\underline{5.7}} & 1.3 & 1.0 & 29.1 & 12.7 & 19.6 & 11.0 & 9.5 & 3.0 & 11.1 \\
 &  & English Words & 9.3 & 3.1 & 0.0 & 1.3 & 12.3 & 1.7 & 16.8 & 8.1 & 6.0 & 0.4 & 5.9 \\
 \cmidrule{2-14}
 & {\multirow{1}{*}{{Caption}}} & Closest Caption & 5.0 & 0.5 & 0.5 & 2.1 & 5.9 & 3.1 & 3.5 & 2.6 & 13.8 & 0.8 & 3.8 \\
\cmidrule{2-14}
& \multicolumn{2}{c|}{\cellcolor{ForestGreen!20}\methodShort} & \textbf{\underline{43.0}}\cellcolor{ForestGreen!20} & 5.1\cellcolor{ForestGreen!20} & 5.1\cellcolor{ForestGreen!20} & \textbf{\underline{6.0}}\cellcolor{ForestGreen!20} & \textbf{\underline{40.2}}\cellcolor{ForestGreen!20} & \textbf{\underline{19.0}}\cellcolor{ForestGreen!20} & \textbf{\underline{36.2}}\cellcolor{ForestGreen!20} & 18.5\cellcolor{ForestGreen!20} & 17.7\cellcolor{ForestGreen!20} & \textbf{\underline{7.9}}\cellcolor{ForestGreen!20} & \textbf{\underline{19.9}}\cellcolor{ForestGreen!20} \\
& \multicolumn{2}{c|}{\cellcolor{ForestGreen!20}\methodEShort} & 41.5\cellcolor{ForestGreen!20} & 5.1\cellcolor{ForestGreen!20} & \textbf{5.8}\cellcolor{ForestGreen!20} & 5.5\cellcolor{ForestGreen!20} & 39.0\cellcolor{ForestGreen!20} & 17.6\cellcolor{ForestGreen!20} & 33.7\cellcolor{ForestGreen!20} & \textbf{\underline{19.2}}\cellcolor{ForestGreen!20} & \textbf{17.8}\cellcolor{ForestGreen!20} & 5.9\cellcolor{ForestGreen!20} & 19.1\cellcolor{ForestGreen!20} \\
\cmidrule{2-14}
& \multicolumn{2}{c|}{\cellcolor{Gray!20}CLIP upper bound}& 93.3\cellcolor{Gray!20} & 60.0\cellcolor{Gray!20} & 47.1\cellcolor{Gray!20} & 47.3\cellcolor{Gray!20} & 88.6\cellcolor{Gray!20} & 94.2\cellcolor{Gray!20} & 95.4\cellcolor{Gray!20} & 70.8\cellcolor{Gray!20} & 95.7\cellcolor{Gray!20} & 81.9\cellcolor{Gray!20} & 77.7\cellcolor{Gray!20} \\
\midrule
& {\multirow{3}{*}{{Caption}}} & BLIP-2 (ViT-L) & 13.4 & 1.4 & 4.8 & 0.0 & 7.5 & 4.7 & 1.7 & 4.7 & 11.6 & 1.1 & 5.1 \\
&  & BLIP-2 (ViT-g) & 16.8 & 1.8 & 4.1 & 0.1 & 13.9 & 7.9 & 2.9 & 5.7 & 24.7 & 1.9 & 8.0 \\
& & LLaVA 1.5 (7B) & 13.8 & 0.9 & 5.1 & 0.0 & 2.9 & 1.7 & 0.3 & 3.5 & 4.6 & 1.4 & 3.4 \\
\midrule
& {\multirow{3}{*}{{VQA}}} & {BLIP-2 (ViT-L)} & 36.1 & 1.8 & 7.0 & 0.1 & 21.5 & 3.7 & 5.7 & 11.5 & 18.9 & 2.5 & 10.9 \\
&  & {BLIP-2 (ViT-g)} & 41.5 & 2.4 & 7.5 & 2.0 & 38.0 & 8.6 & 10.2 & 13.8 & 33.2 & 2.8 & 16.0  \\
& & LLaVA 1.5 (7B) & 41.4 & 0.4 & 6.1 & 0.0 & 5.0 & 3.0 & 0.7 & 6.6 & 1.3 & 2.0 & 6.6 \\
\bottomrule
\end{tabular}
\vspace{2pt}
\caption{Semantic IoU on the ten datasets. \inlineColorbox{ForestGreen!20}{Green} is our method, \inlineColorbox{Gray!20}{gray} shows the upper bound. \textbf{Bold} represents best, \underline{underline} indicates best considering also image captioning and VQA models.}
\label{tab:extended_metric_semantic_iou}
\end{table*}

\begin{table*}[t]
\small
\begin{tabular}{ccc|cccccccccc|c}
\toprule
& \multicolumn{2}{c|}{\textbf{Method}}  &\multicolumn{11}{c}{\textbf{Semantic Similarity (x100)} $\uparrow$} \\
& & & {C101} & {DTD} & {ESAT} & Airc. & Flwr & Food & {Pets} & {SUN} & {Cars} & {UCF} & \textbf{Avg.} \\
\midrule
\scriptsize \parbox[t]{2mm}{\multirow{6}{*}{\rotatebox{90}{CLIP RN50}}} & {\multirow{2}{*}{{CLIP}}} & WordNet & 43.2 & 29.0 & 18.5 & 21.6 & 46.7 & 44.6 & 50.3 & 42.8 & 26.4 & 40.0 & 36.3 \\
& & English Words & 36.0 & 29.5 & 14.9 & 20.0 & 38.1 & 34.2 & 40.7 & 35.4 & 18.3 & 32.4 & 29.9 \\
\cmidrule{2-14}
& {\multirow{1}{*}{{Caption}}} & Closest Caption & 37.2 & 22.8 & 14.2 & 26.8 & 38.9 & 41.2 & 32.6 & 37.4 & \textbf{44.3} & 32.4 & 32.8 \\
\cmidrule{2-14}
& \multicolumn{2}{c|}{\cellcolor{ForestGreen!20}\methodShort} & \textbf{62.3}\cellcolor{ForestGreen!20} & 36.4\cellcolor{ForestGreen!20} & 22.6\cellcolor{ForestGreen!20} & \textbf{28.7}\cellcolor{ForestGreen!20} & \textbf{52.8}\cellcolor{ForestGreen!20} & \textbf{\underline{59.0}}\cellcolor{ForestGreen!20} & 57.0\cellcolor{ForestGreen!20} & \textbf{50.2}\cellcolor{ForestGreen!20} & 42.9\cellcolor{ForestGreen!20} & \textbf{46.2}\cellcolor{ForestGreen!20} & \textbf{45.8}\cellcolor{ForestGreen!20} \\
& \multicolumn{2}{c|}{\cellcolor{ForestGreen!20}\methodEShort} & \textbf{62.3}\cellcolor{ForestGreen!20} & \textbf{\underline{37.8}}\cellcolor{ForestGreen!20} & \textbf{23.6}\cellcolor{ForestGreen!20} & 26.0\cellcolor{ForestGreen!20} & 52.4\cellcolor{ForestGreen!20} & 58.3\cellcolor{ForestGreen!20} & \textbf{\underline{57.9}}\cellcolor{ForestGreen!20} & \textbf{50.2}\cellcolor{ForestGreen!20} & 41.6\cellcolor{ForestGreen!20} & 46.1\cellcolor{ForestGreen!20} & 45.6\cellcolor{ForestGreen!20} \\
\cmidrule{2-14}
& \multicolumn{2}{c|}{\cellcolor{Gray!20}CLIP upper bound} & 88.1\cellcolor{Gray!20} & 62.4\cellcolor{Gray!20} & 52.8\cellcolor{Gray!20} & 53.9\cellcolor{Gray!20} & 72.0\cellcolor{Gray!20} & 83.7\cellcolor{Gray!20} & 85.8\cellcolor{Gray!20} & 73.9\cellcolor{Gray!20} & 81.0\cellcolor{Gray!20} & 73.9\cellcolor{Gray!20} & 72.7\cellcolor{Gray!20} \\
\midrule
\scriptsize \parbox[t]{2mm}{\multirow{6}{*}{\rotatebox{90}{CLIP ViT-L/14}}} & {\multirow{2}{*}{{CLIP}}} & WordNet & 48.6 & 32.7 & 24.4 & 18.9 & 55.9 & 49.6 & 53.7 & 44.9 & 28.8 & 44.2 & 40.2 \\
& & English Words & 39.3 & 31.6 & 19.1 & 18.6 & 43.4 & 38.0 & 44.2 & 36.0 & 19.9 & 34.7 & 32.5 \\
\cmidrule{2-14}
& {\multirow{1}{*}{{Caption}}} & Closest Caption & 42.1 & 23.9 & 23.4 & 29.2 & 40.0 & 46.9 & 40.2 & 39.8 & \textbf{49.2} & 40.3 & 37.5 \\
\cmidrule{2-14}
& \multicolumn{2}{c|}{\cellcolor{ForestGreen!20}\methodShort} & 65.7\cellcolor{ForestGreen!20} & 40.0\cellcolor{ForestGreen!20} & 32.0\cellcolor{ForestGreen!20} & \textbf{30.3}\cellcolor{ForestGreen!20} & 55.5\cellcolor{ForestGreen!20} & 64.5\cellcolor{ForestGreen!20} & 62.5\cellcolor{ForestGreen!20} & 52.5\cellcolor{ForestGreen!20} & 47.4\cellcolor{ForestGreen!20} & \textbf{54.1}\cellcolor{ForestGreen!20} & 50.4\cellcolor{ForestGreen!20} \\
& \multicolumn{2}{c|}{\cellcolor{ForestGreen!20}\methodEShort} & \textbf{66.3}\cellcolor{ForestGreen!20} & \textbf{\underline{40.3}}\cellcolor{ForestGreen!20} & \textbf{34.9}\cellcolor{ForestGreen!20} & 27.0\cellcolor{ForestGreen!20} & \textbf{56.1}\cellcolor{ForestGreen!20} & \textbf{\underline{65.0}}\cellcolor{ForestGreen!20} & \textbf{\underline{62.9}}\cellcolor{ForestGreen!20} & \textbf{53.0}\cellcolor{ForestGreen!20} & 46.5\cellcolor{ForestGreen!20} & 53.8\cellcolor{ForestGreen!20} & \textbf{\underline{50.6}}\cellcolor{ForestGreen!20} \\
\cmidrule{2-14}
& \multicolumn{2}{c|}{\cellcolor{Gray!20}CLIP upper bound} & 90.8\cellcolor{Gray!20} & 69.8\cellcolor{Gray!20} & 67.7\cellcolor{Gray!20} & 66.7\cellcolor{Gray!20} & 83.4\cellcolor{Gray!20} & 93.7\cellcolor{Gray!20} & 91.8\cellcolor{Gray!20} & 80.5\cellcolor{Gray!20} & 92.3\cellcolor{Gray!20} & 83.3\cellcolor{Gray!20} & 82.0\cellcolor{Gray!20}  \\
\midrule
\scriptsize\parbox[t]{2mm}{\multirow{7}{*}{\rotatebox{90}{SigLIP ViT-L/16}}} & {\multirow{2}{*}{{SigLIP}}} & WordNet & 50.7 & 40.2 & 19.1 & 22.0 & 53.3 & 56.2 & 60.2 & 46.7 & 33.7 & 43.2 & 42.5 \\
 &  & English Words & 40.0 & 36.2 & 18.8 & 18.6 & 40.1 & 40.7 & 49.2 & 36.1 & 28.1 & 34.1 & 34.2 \\
 \cmidrule{2-14}
 & {\multirow{1}{*}{{Caption}}} & Closest Caption & 45.0 & 24.1 & 17.2 & 30.6 & 41.1 & 49.1 & 41.3 & 41.2 & \textbf{53.3} & 40.1 & 38.3 \\
\cmidrule{2-14}
& \multicolumn{2}{c|}{\cellcolor{ForestGreen!20}\methodShort} & \textbf{71.3}\cellcolor{ForestGreen!20} & \textbf{\underline{41.6}}\cellcolor{ForestGreen!20} & 28.0\cellcolor{ForestGreen!20} & \textbf{\underline{34.4}}\cellcolor{ForestGreen!20} & \textbf{\underline{61.7}}\cellcolor{ForestGreen!20} & \textbf{\underline{66.2}}\cellcolor{ForestGreen!20} & \textbf{\underline{66.0}}\cellcolor{ForestGreen!20} & 54.2\cellcolor{ForestGreen!20} & 48.4\cellcolor{ForestGreen!20} & \textbf{\underline{55.1}}\cellcolor{ForestGreen!20} & \textbf{\underline{52.7}}\cellcolor{ForestGreen!20} \\
& \multicolumn{2}{c|}{\cellcolor{ForestGreen!20}\methodEShort} & 70.7\cellcolor{ForestGreen!20} & 41.0\cellcolor{ForestGreen!20} & \textbf{33.6}\cellcolor{ForestGreen!20} & 27.7\cellcolor{ForestGreen!20} & 58.2\cellcolor{ForestGreen!20} & 61.7\cellcolor{ForestGreen!20} & 63.9\cellcolor{ForestGreen!20} & \textbf{\underline{54.3}}\cellcolor{ForestGreen!20} & 48.5\cellcolor{ForestGreen!20} & 54.2\cellcolor{ForestGreen!20} & 51.4\cellcolor{ForestGreen!20} \\
\cmidrule{2-14}
& \multicolumn{2}{c|}{\cellcolor{Gray!20}CLIP upper bound}& 97.8\cellcolor{Gray!20} & 75.6\cellcolor{Gray!20} & 63.1\cellcolor{Gray!20} & 80.0\cellcolor{Gray!20} & 92.1\cellcolor{Gray!20} & 96.4\cellcolor{Gray!20} & 96.8\cellcolor{Gray!20} & 83.2\cellcolor{Gray!20} & 98.1\cellcolor{Gray!20} & 89.6\cellcolor{Gray!20} & 87.3\cellcolor{Gray!20} \\
\midrule
& {\multirow{3}{*}{{Caption}}} & BLIP-2 (ViT-L) & 57.8 & 31.4 & 39.9 & 24.4 & 36.1 & 44.6 & 29.0 & 45.3 & 46.4 & 38.0 & 39.3 \\
&  & BLIP-2 (ViT-g) & 63.0 & 33.1 & 36.2 & 24.3 & 45.2 & 51.6 & 31.6 & 48.3 & 61.0 & 44.6 & 43.9 \\
& & LLaVA 1.5 (7B) &  56.8 & 29.4 & 40.5 & 21.3 & 31.1 & 36.9 & 24.8 & 42.5 & 38.1 & 37.9 & 35.9 \\
\midrule
& {\multirow{3}{*}{{VQA}}} & {BLIP-2 (ViT-L)} & 70.5 & 34.9 & 29.7 & 29.1 & 48.8 & 42.0 & 40.0 & 50.6 & 52.4 & 48.6 & 44.7 \\
&  & {BLIP-2 (ViT-g)} & 73.5 & 36.5 & 31.4 & 30.8 & 59.9 & 52.1 & 43.9 & 53.3 & 65.1 & 55.1 & 50.1 \\
& & LLaVA 1.5 (7B) & 72.6 & 36.7 & 44.1 & 29.4 & 41.8 & 41.1 & 36.0 & 41.9 & 35.3 & 46.6 & 42.6 \\
\bottomrule
\end{tabular}
\vspace{2pt}
\caption{Semantic similarity on the ten datasets. \inlineColorbox{ForestGreen!20}{Green} is ours, \inlineColorbox{Gray!20}{gray} shows the upper bound. \textbf{Bold} represents best, \underline{underline} indicates best considering also image captioning and VQA models.}
\label{tab:extended_metric_semantic_similarity}
\end{table*}

\section*{E. Backbone architecture}
\label{sec:ablation_backbone}

To answer the natural question about whether the outcome of our model depends on the backbone architecture, we further extend our main results with a CLIP model with a ResNet50 architecture and with a ViT-L/16 architecture pre-trained with the sigmoid loss~\cite{zhai2023sigmoid}.
We report this additional ablation in Tab.~\ref{tab:extended_metric_semantic_cluster_acc}, Tab.~\ref{tab:extended_metric_semantic_iou}, and Tab.~\ref{tab:extended_metric_semantic_similarity} for the cluster accuracy, the semantic IoU, and the semantic similarity, respectively.
We can see that the performance with the CLIP ResNet50 is lower across all the metrics compared to CLIP ViT-L/14.
This is expected since ResNet50 is a a smaller architecture, thus with a reduced capacity for semantic representation learning as compared to ViT-L/14.
Nevertheless, our method with ResNet50 is still competitive against BLIP-2 models while using 40x fewer parameters (note that our ViT-L implementation uses 10x fewer parameters). A similar trend occurs for the SigLIP family of models, as their performance generally surpasses CLIP with the ViT-L/14 backbone, while only having approximately $1.5\times$ more parameters. Notably, performance improves consistently, with an average improvement of +$3.8\%$ for cluster accuracy, $+1.6$ for semantic IoU, and $+2.1$ for semantic similarity. Differently from the smaller models, we do not experience a consistent improvement of \methodEShort\ against \methodShort\ with the SigLIP pre-trained backbone, where we notice a negative fluctuation on the three metrics when applying prompt ensembling.

\section*{F. Qualitative results}
\label{sec:qualitatives}

\begin{figure*}
\centering
\begin{tabular}{lllll}
{\includegraphics[width=2.5cm]{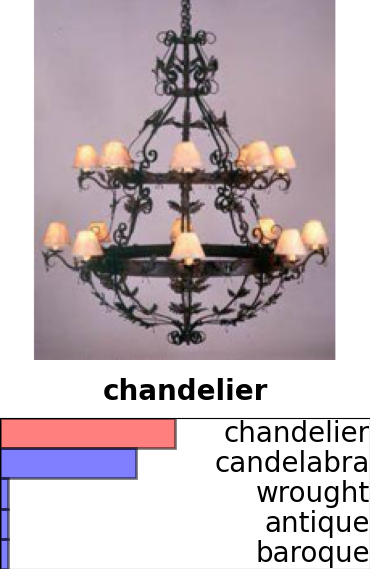}} &
{\includegraphics[width=2.5cm]{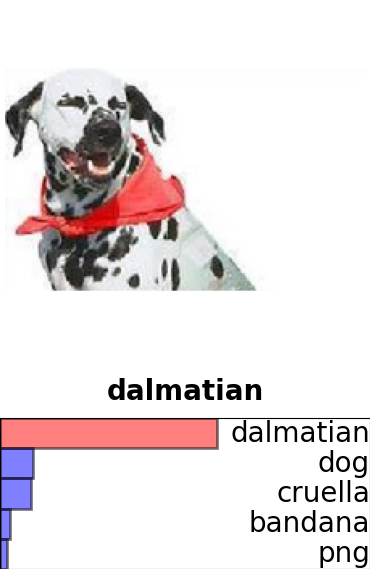}} &
{\includegraphics[width=2.5cm]{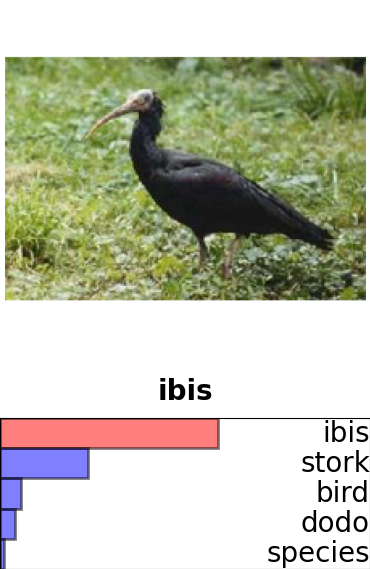}} &
{\includegraphics[width=2.5cm]{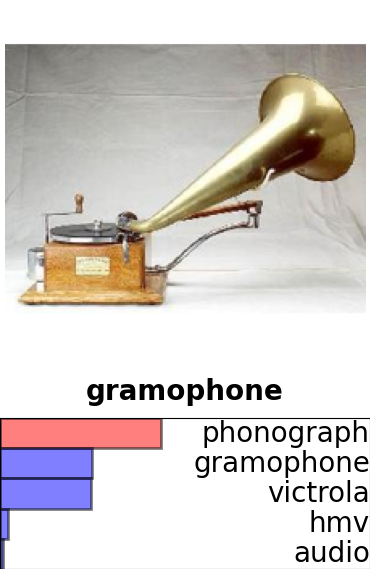}} &
{\includegraphics[width=2.5cm]{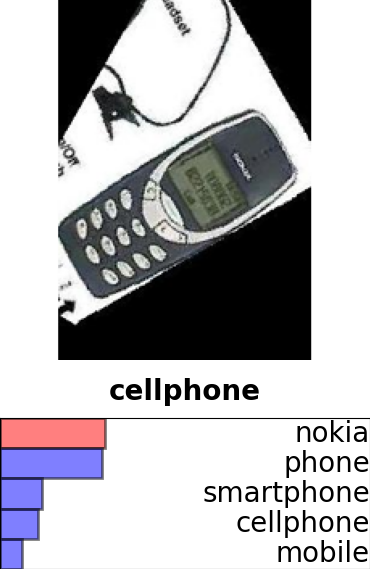}} \\
\end{tabular}
\caption{Qualitative results of \methodShort\ on Caltech-101. The first three samples represent success cases, the last two shows failure cases.}\vspace{-2pt}
\label{fig:qualitative_1}
\end{figure*}

\begin{figure*}
\centering
\begin{tabular}{lllll}
{\includegraphics[width=2.5cm]{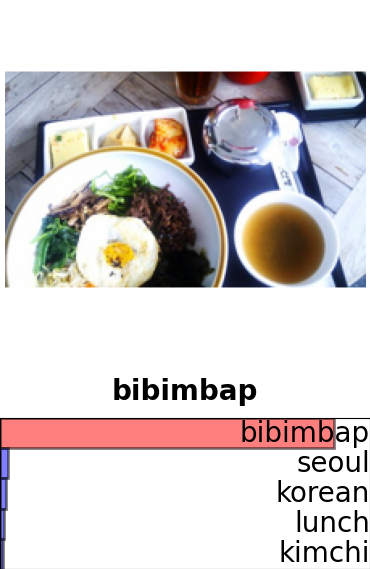}} &
{\includegraphics[width=2.5cm]{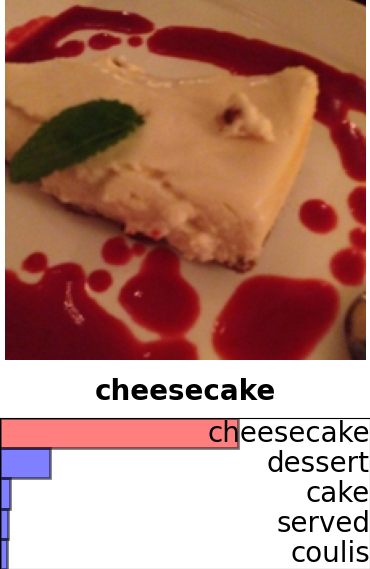}} &
{\includegraphics[width=2.5cm]{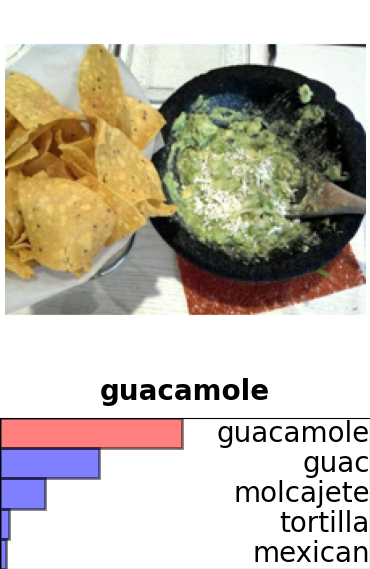}} &
{\includegraphics[width=2.5cm]{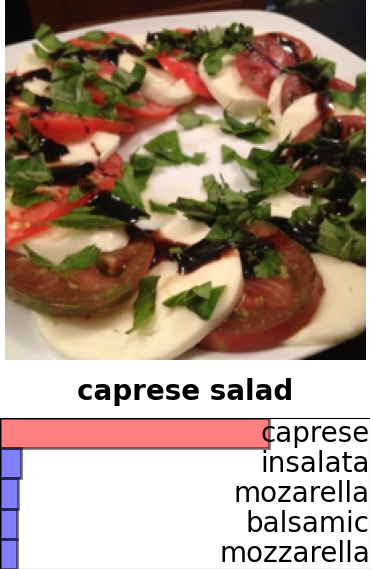}} &
{\includegraphics[width=2.5cm]{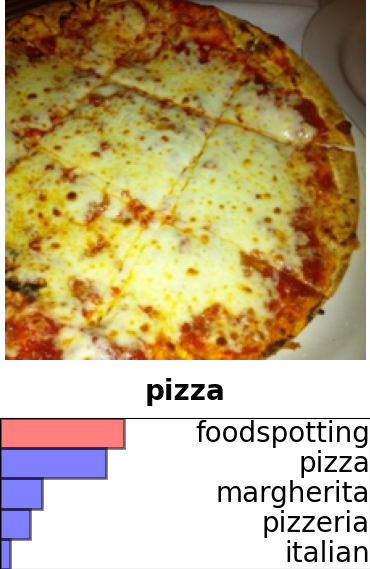}} \\
\end{tabular}
\caption{Qualitative results of \methodShort\ on Food101. The first three samples represent success cases, the last two shows failure cases.}\vspace{-2pt}
\label{fig:qualitative_2}
\end{figure*}

\begin{figure*}
\centering
\begin{tabular}{lllll}
{\includegraphics[width=2.5cm]{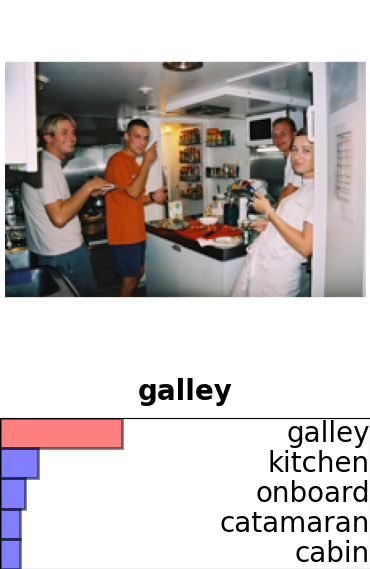}} &
{\includegraphics[width=2.5cm]{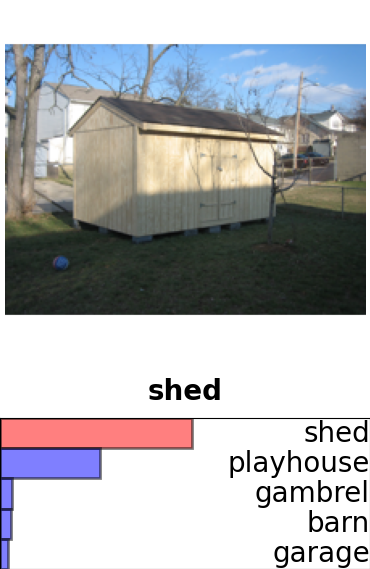}} &
{\includegraphics[width=2.5cm]{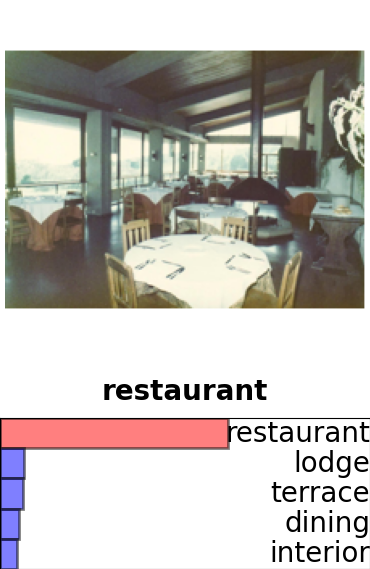}} &
{\includegraphics[width=2.5cm]{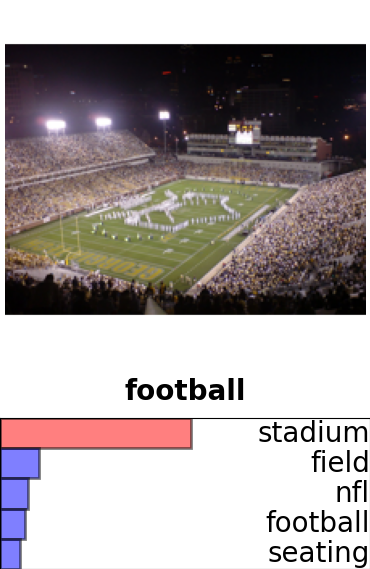}} &
{\includegraphics[width=2.5cm]{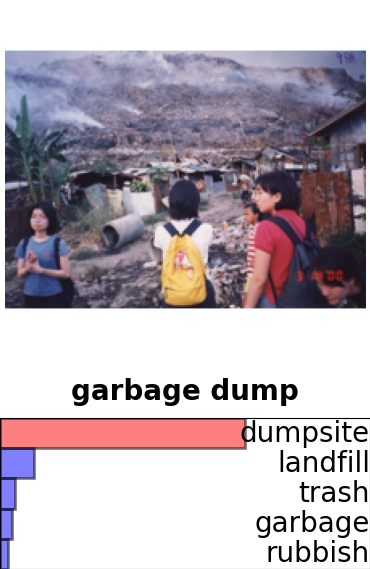}} \\
\end{tabular}
\caption{Qualitative results of \methodShort\ on SUN397. The first three samples represent success cases, the last two shows failure cases.}
\label{fig:qualitative_3}
\end{figure*}

\noindent\textbf{Classification.} Last, we report some qualitative results of \methodShort\ applied on three different datasets, namely Caltech-101~(Fig.~\ref{fig:qualitative_1}), Food101~(Fig.~\ref{fig:qualitative_2}), and SUN397~(Fig.~\ref{fig:qualitative_3}), where the first is coarse, and the last two are fine-grained, focusing on food plates and places respectively.
For each, we present a batch of five images, where the first three represent success cases and the last two show interesting failure cases.
Each sample shows the image we input to our method with the top-5 candidate classes.

From the results, we can see that for many success cases, our method not only generates the correct class name and selects it as the best matching label, but it also provides valid alternatives for classification.
For example, the third image in Fig~\ref{fig:qualitative_1} or the second image in Fig~\ref{fig:qualitative_2}, where \methodShort\ provides the names "dessert" for the cheesecake and the label "bird" for the ibis.
This phenomenon also happens in failure cases, where \eg~the last sample in Fig~\ref{fig:qualitative_2} provides both the name "pizza" and the name "margherita" for the dish, despite selecting the wrong name from the set.

Another interesting observation is that our method provides names for different objects in the same scene.
For instance, the third and fourth samples in Fig~\ref{fig:qualitative_2} contain labels for both "guacamole" and  "tortillas" for the first, and for "mozzarella", "insalata", and "balsamic" for the second.
A further detail on the latter case is the ability of CLIP to reason in multiple languages since "insalata" translates to "salad" from Italian to English.

Regarding failure cases, it is interesting to note that the candidate names and the predicted label often describe well the input image despite being wrong w.r.t~the dataset label.
For instance, the two failure cases in Fig.~\ref{fig:qualitative_3} select "stadium" and "dumpsite" when the ground-truth class names are "football" and "garbage site".
In addition, for the first case, the exact name "football" is still available among the best candidate names, but our method considers "stadium" as a better fit.
Another example is the last failure case in Fig~\ref{fig:qualitative_1}, where the model assigns the name "nokia" to a Nokia 3310 cellphone, while the ground-truth class name is "cellphone".
Also in this case, the ground-truth label is present in the candidate list but our method considers "nokia" a more fitting class.

Finally, we notice the discovery of correlations between terms in the reasoning of our model.
In the provided examples, it happens multiple times that the candidate class names do not describe objects in the scene but rather a correlated concept to the image.
For instance, the third example in Fig~\ref{fig:qualitative_1} shows a Dalmatian, and among the candidate names there is "cruella", which is the name of the villain of the movie "The Hundred and One Dalmatians".
Another instance of this appears in the first example of Fig~\ref{fig:qualitative_2}, where the model correctly associates the "bibimbap" dish to its place of origin, Korea, with the candidate name "korean".

\begin{figure*}
\centering
\begin{tabular}{ll}
{\includegraphics[width=0.45\linewidth]{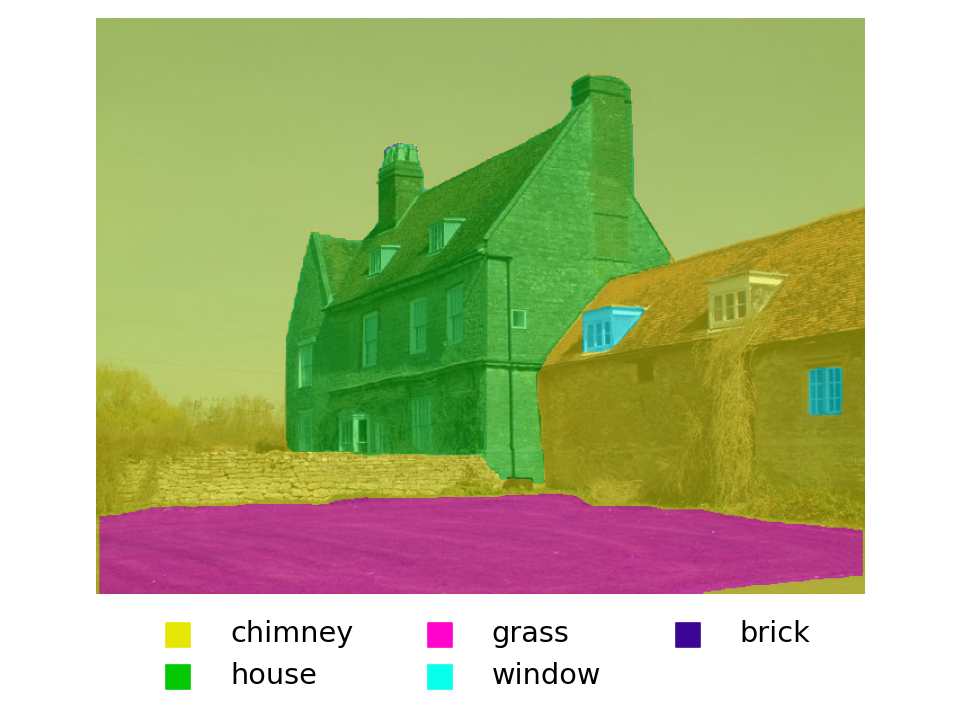}} &
{\includegraphics[width=0.45\linewidth]{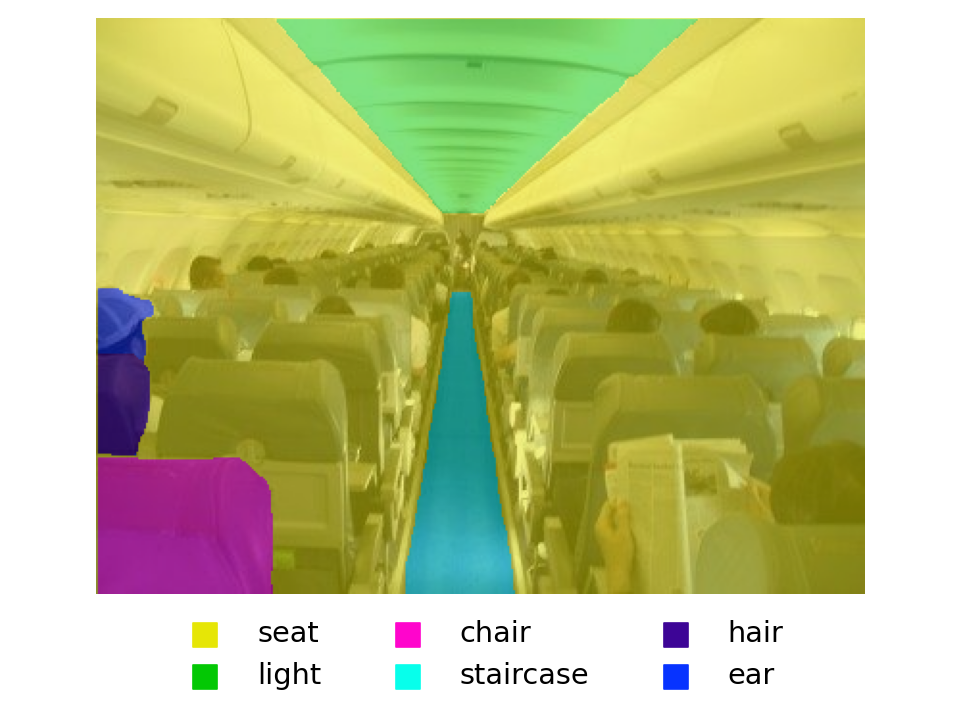}} \\
\multicolumn{2}{c}{(a) SAM + LLaVA 1.5 (7B)} \\
{\includegraphics[width=0.45\linewidth]{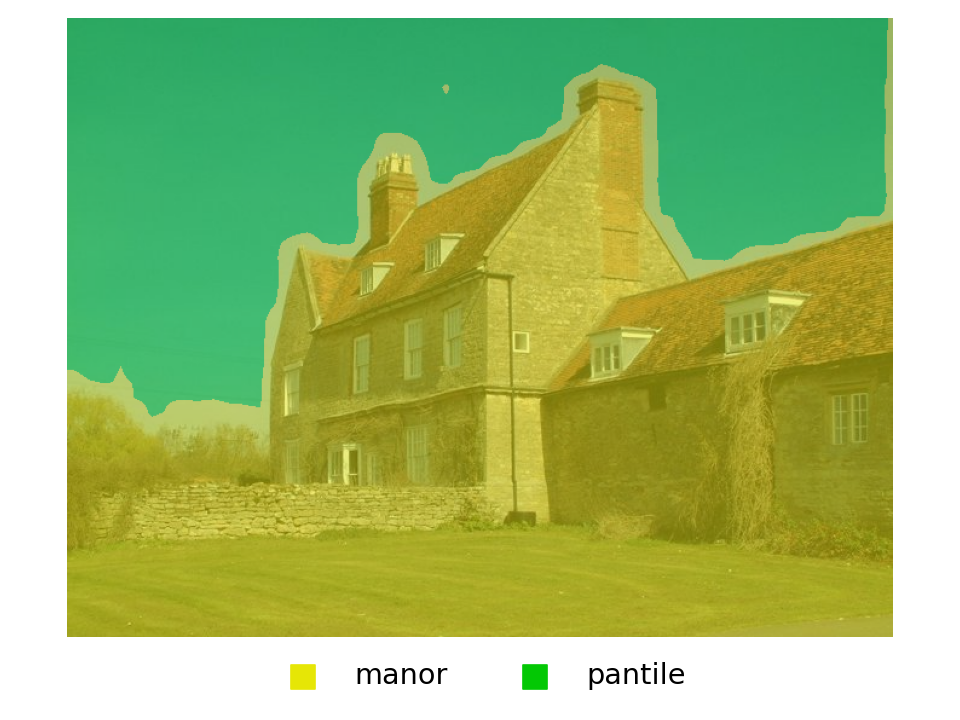}} &
{\includegraphics[width=0.45\linewidth]{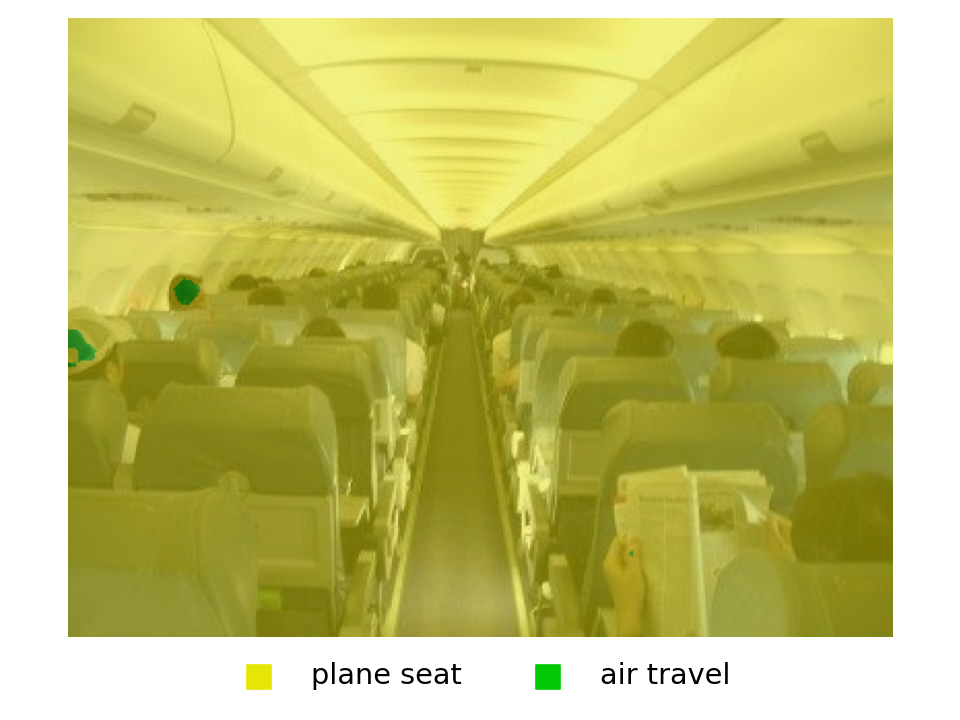}} \\
\multicolumn{2}{c}{(b) SAN + WordNet} \\
{\includegraphics[width=0.45\linewidth]{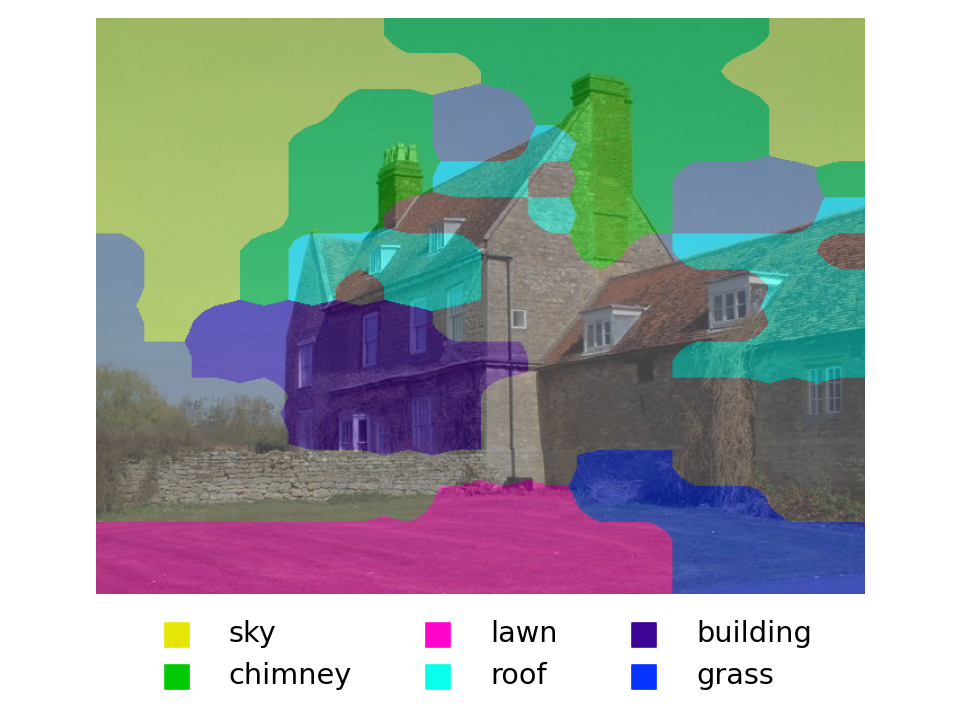}} &
{\includegraphics[width=0.45\linewidth]{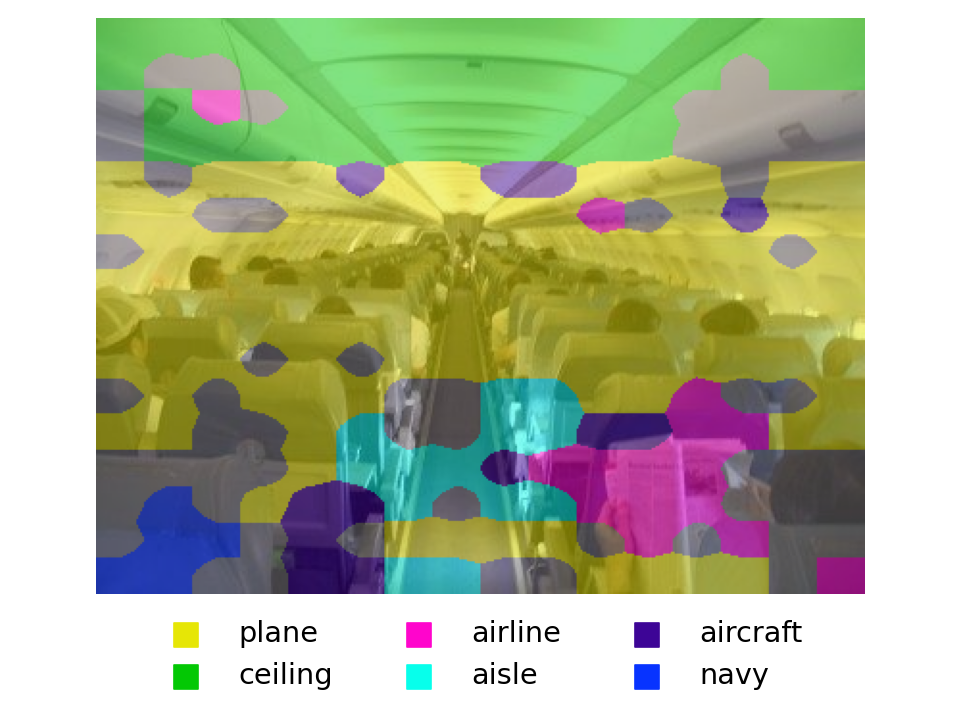}} \\
\multicolumn{2}{c}{(c) \methodSShort} \\
\end{tabular}
\caption{Qualitative results of \taskSName\ methods on ADE20K-150. For visualisation purposes, we visualise only the top-6 most prominent regions.}
\label{fig:qualitative_4}
\end{figure*}

\begin{figure*}
\centering
\begin{tabular}{ll}
{\includegraphics[width=0.45\linewidth]{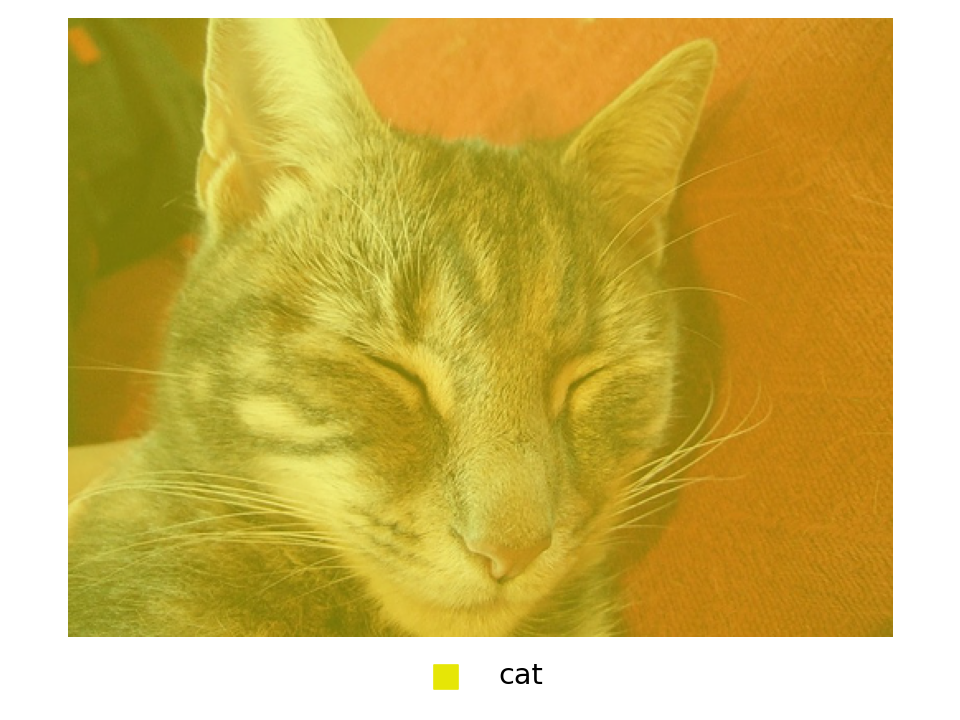}} &
{\includegraphics[width=0.45\linewidth]{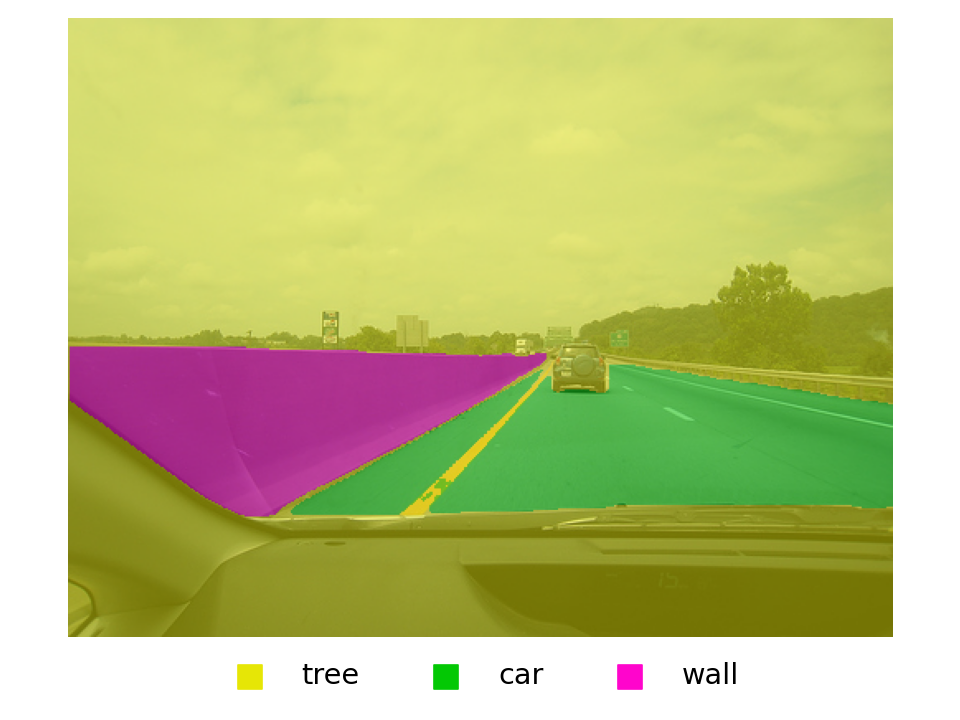}} \\
\multicolumn{2}{c}{(a) SAM + LLaVA 1.5 (7B)} \\
{\includegraphics[width=0.45\linewidth]{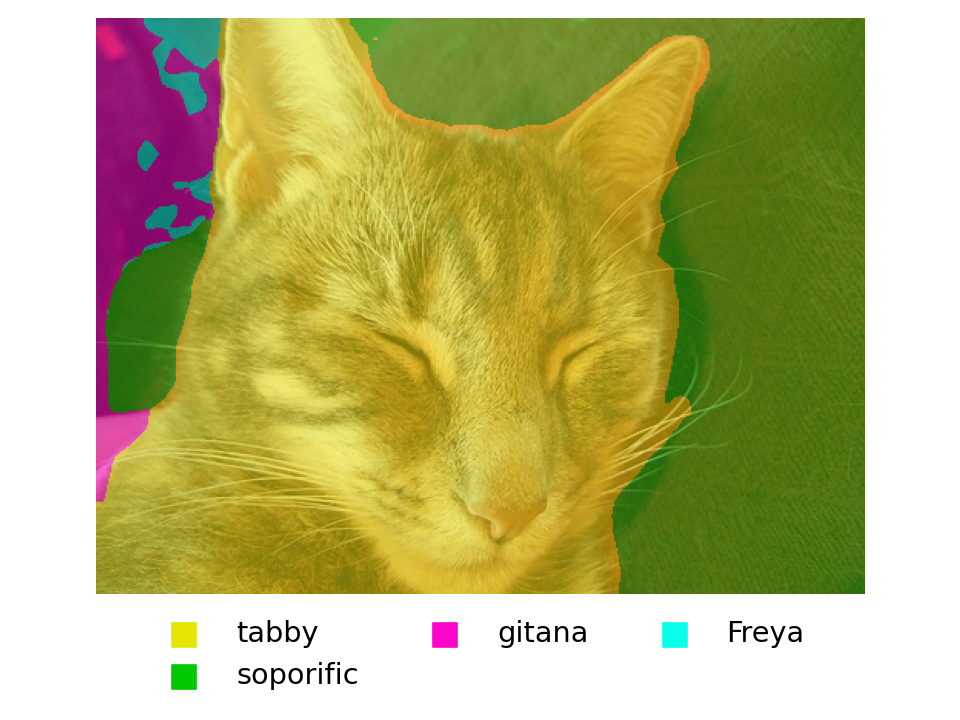}} &
{\includegraphics[width=0.45\linewidth]{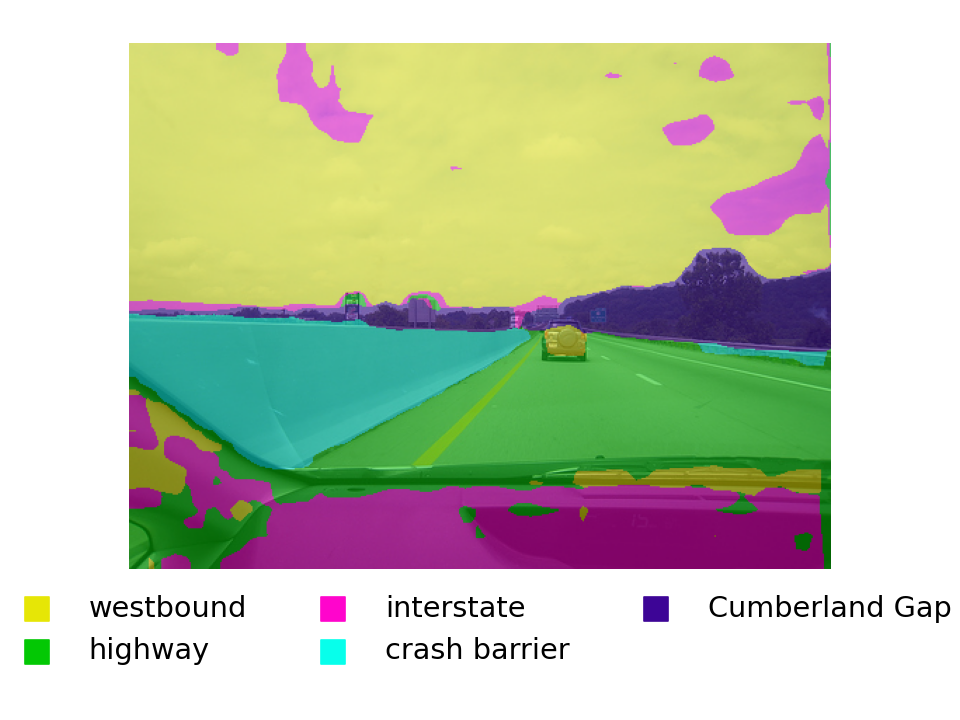}} \\
\multicolumn{2}{c}{(b) SAN + WordNet} \\
{\includegraphics[width=0.45\linewidth]{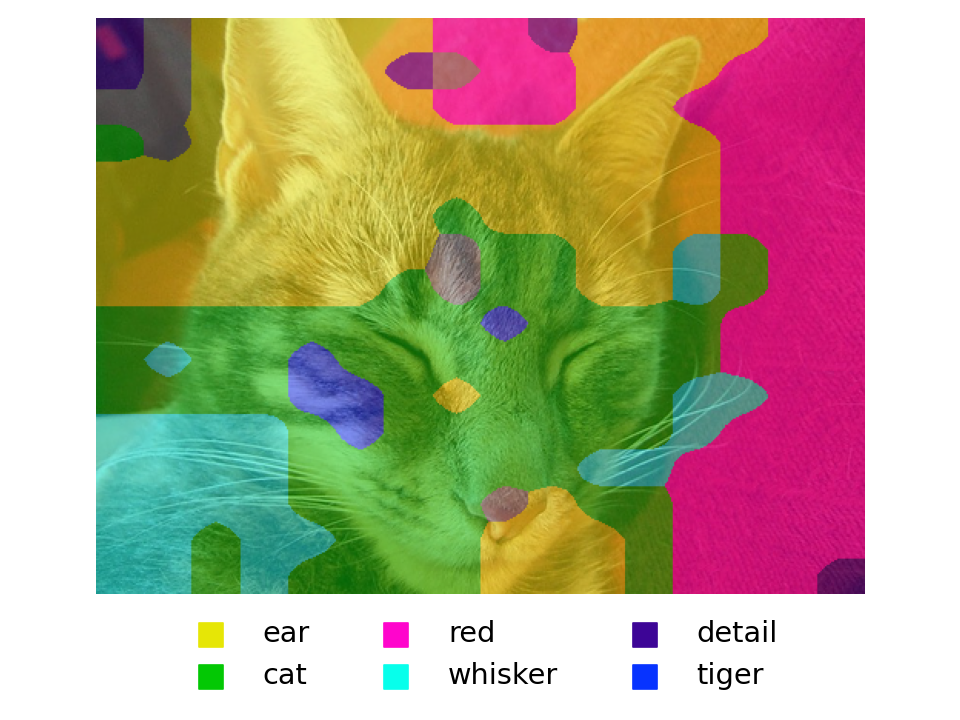}} &
{\includegraphics[width=0.45\linewidth]{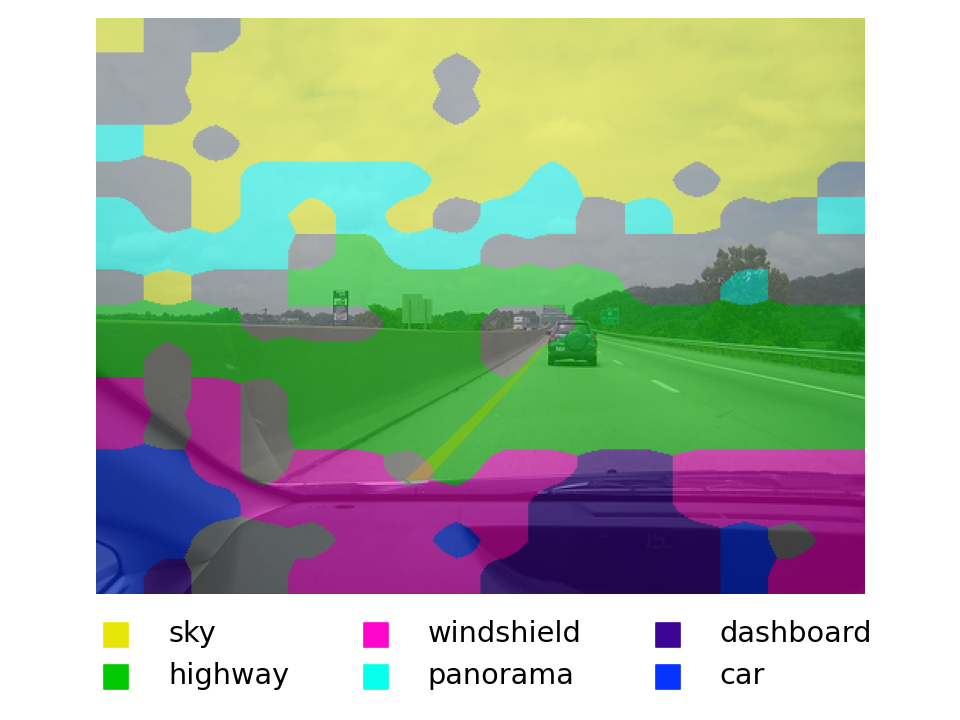}} \\
\multicolumn{2}{c}{(c) \methodSShort} \\
\end{tabular}
\caption{Qualitative results of \taskSName\ methods on PASCAL Context-59. For visualisation purposes, we visualise only the top-6 most prominent regions.}
\label{fig:qualitative_5}
\end{figure*}

\begin{figure*}
\centering
\begin{tabular}{ll}
{\includegraphics[width=0.45\linewidth]{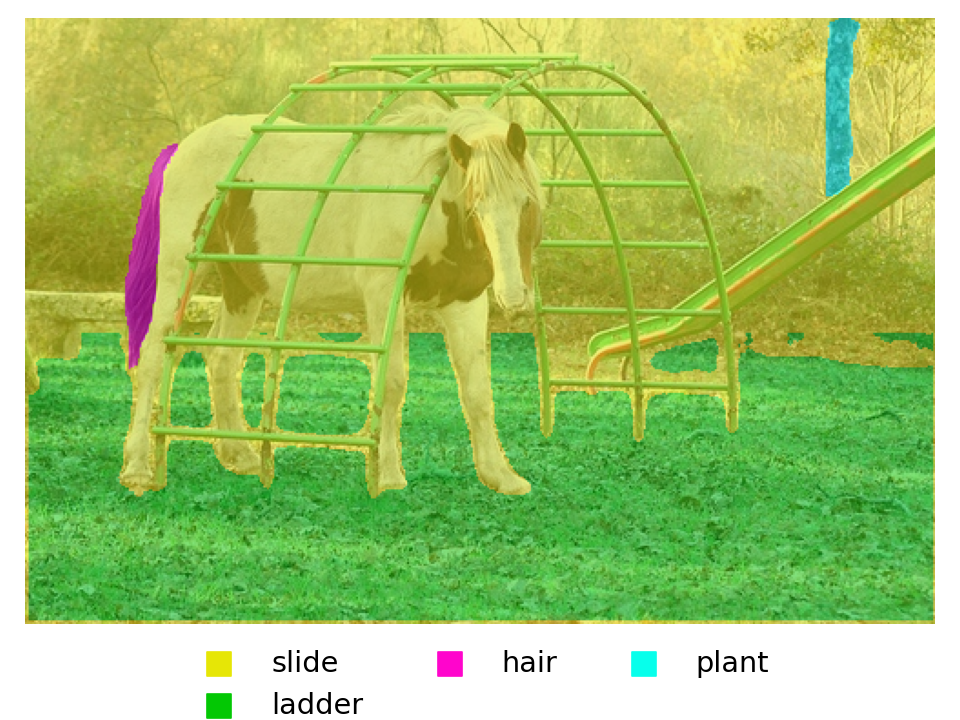}} &
{\includegraphics[width=0.45\linewidth]{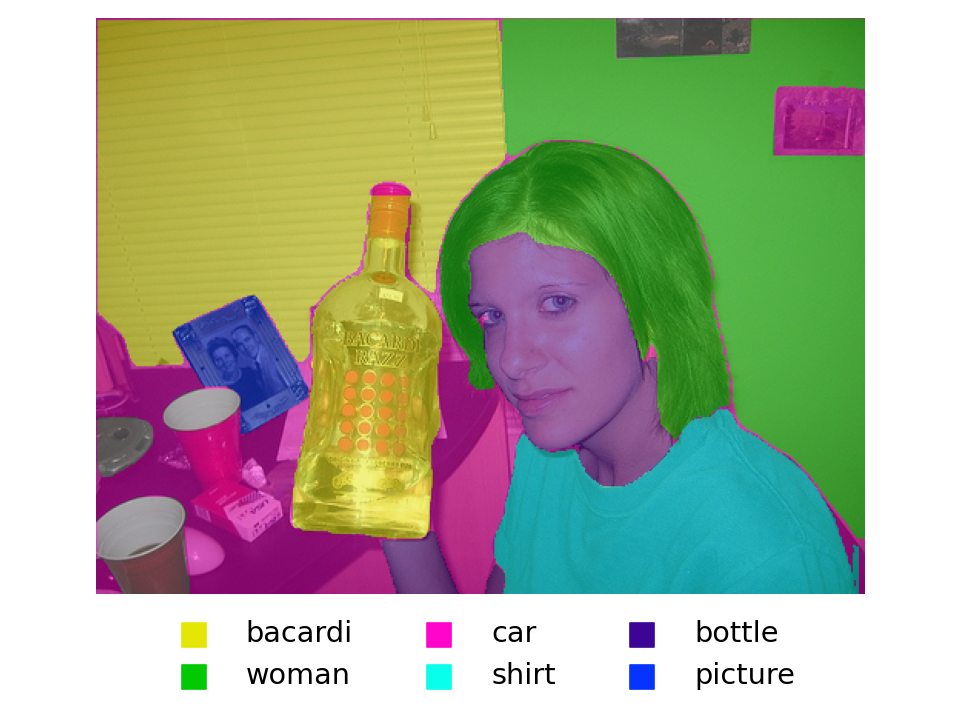}} \\
\multicolumn{2}{c}{(a) SAM + LLaVA 1.5 (7B)} \\
{\includegraphics[width=0.45\linewidth]{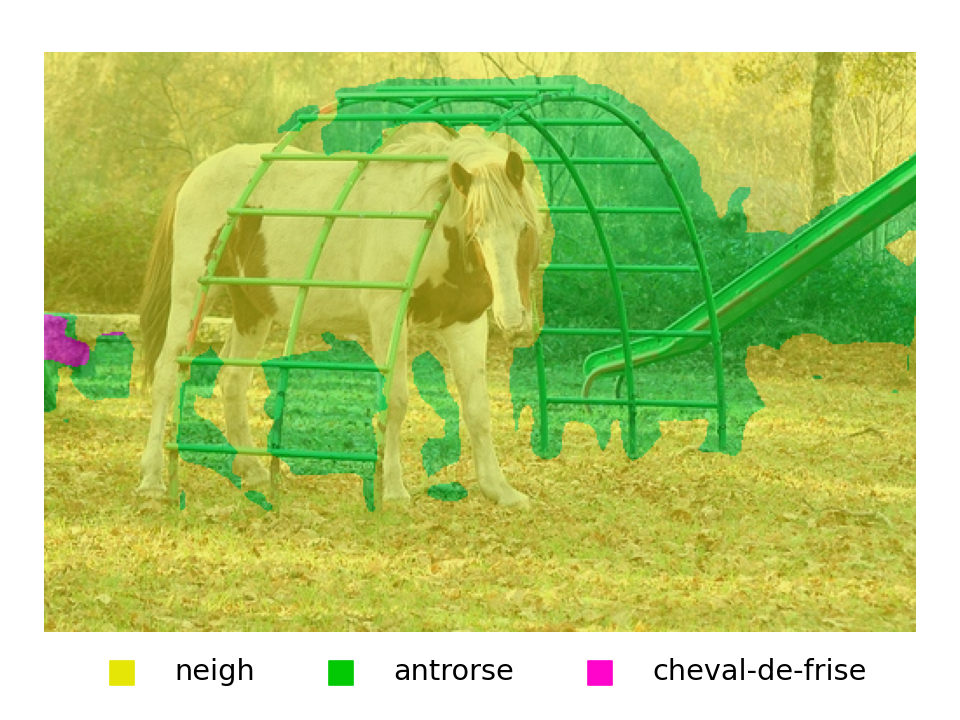}} &
{\includegraphics[width=0.45\linewidth]{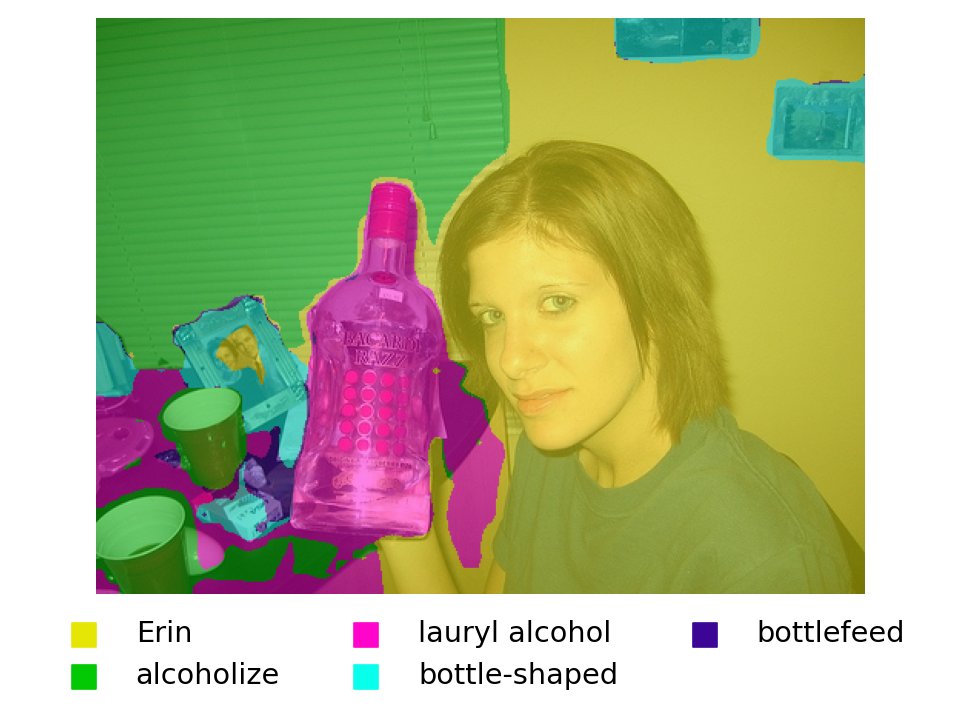}} \\
\multicolumn{2}{c}{(b) SAN + WordNet} \\
{\includegraphics[width=0.45\linewidth]{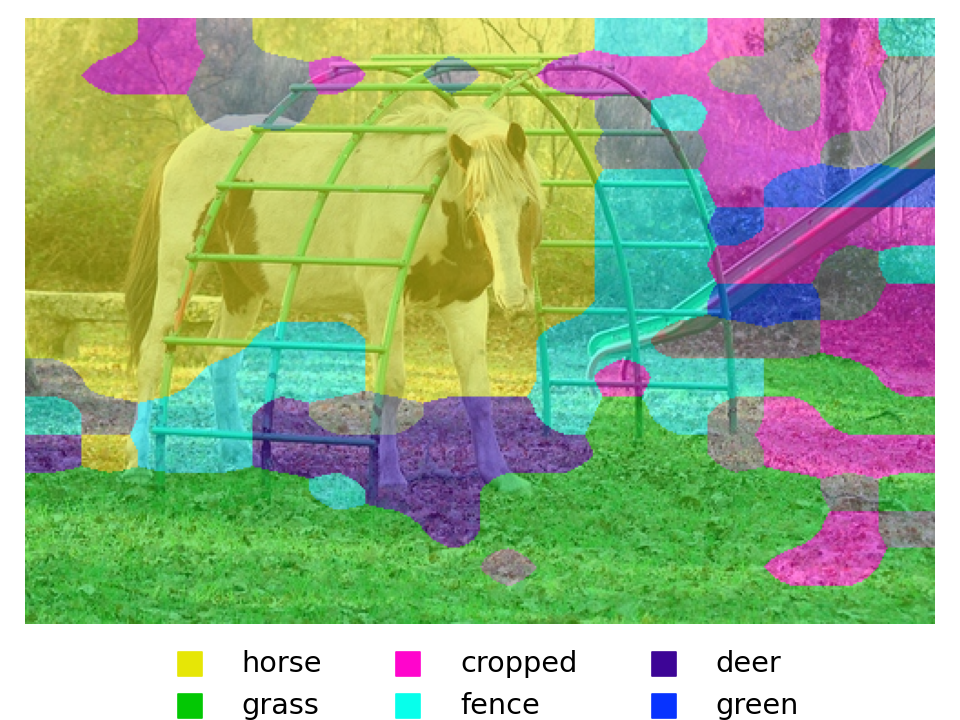}} &
{\includegraphics[width=0.45\linewidth]{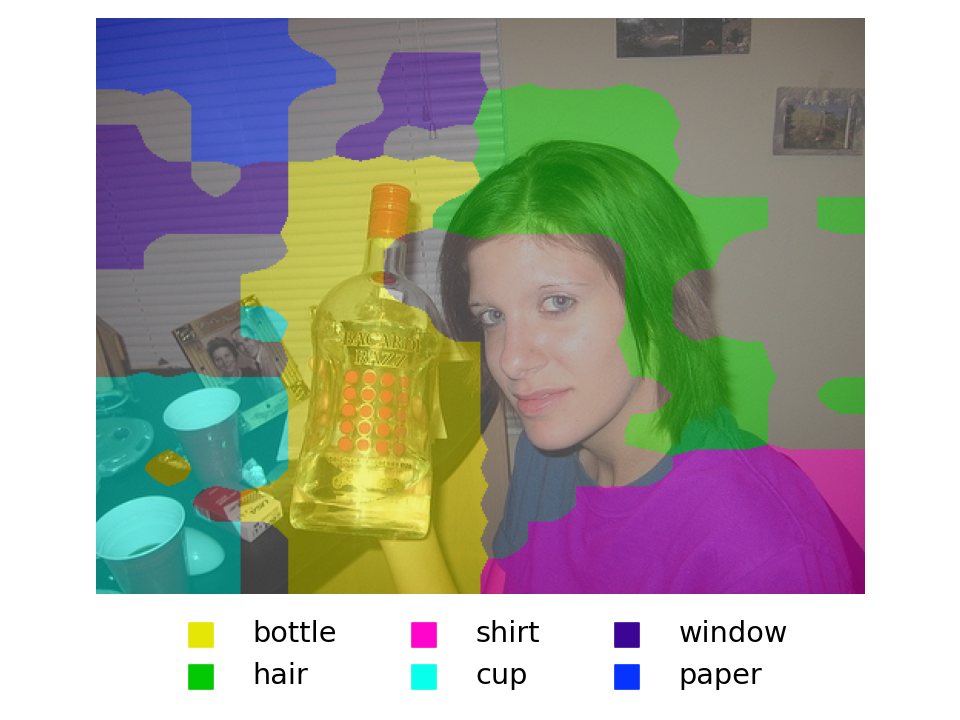}} \\
\multicolumn{2}{c}{(c) \methodSShort} \\
\end{tabular}
\caption{Qualitative results of \taskSName\ methods on PascalVOC-20. For visualisation purposes, we visualise only the top-6 most prominent regions.}
\label{fig:qualitative_6}
\end{figure*}

\noindent\textbf{Segmentation.} In a similar fashion, we report some qualitative results of \methodShort\ for semantic segmentation, {taking the best models in each group: of SAM with LLaVA 1.5 (7B), SAN with WordNet as vocabulary, and our localizer-free model for \taskSShort, \methodSShort}. We report some output examples in Fig.~\ref{fig:qualitative_4}, Fig.~\ref{fig:qualitative_5}, and Fig.~\ref{fig:qualitative_6}. From a segmentation perspective, it is notable how SAM achieves the best segmentation masks, despite missing some more local objects, \eg, it misses all the objects except the cat in the first example of Fig.~\ref{fig:qualitative_5}, or the horse in the first example of Fig.~\ref{fig:qualitative_6}. Moreover, it sometimes misclassifies one object for another (\eg, it considers the table as a "car" in the second example of Fig.~\ref{fig:qualitative_6}, or the aisle in the airplane as "staircase" in the second example of Fig.~\ref{fig:qualitative_4}, or assigns the label "tree" to the sky and the car in Fig.~\ref{fig:qualitative_5}). In some other cases, the most correct label for a region is assigned to a near segmentation mask, for instance in the second examples in Fig.~\ref{fig:qualitative_6} where it assigns "woman" to the hair and the wall behind the person, or "bottle" to the arm holding the bottle in the same example. {We highlight that, nevertheless, SAM + LLaVA 1.5 (7B) is the model with the highest number of parameters overall, using two large-scale foundation models to address \taskSShort.}

Differently from SAM, SAN tends to segment whole objects instead of parts, \eg, first and second example in Fig.~\ref{fig:qualitative_4}, where it divides the image into foreground and background, or completely merge floor and ceiling of the airplane. Moreover, given the labels selected by SAN, it looks like the model tend to assign the most out-of-context words compared with SAM with LLaVA 1.5 (7B) and \methodSShort. As an example, SAN assigns the word "Erin" to the person in the second example of Fig.~\ref{fig:qualitative_6}, or "Cumberland Gap" to the landline in the second example in Fig.~\ref{fig:qualitative_5}. Other examples include "pantile" or "westbound" assigned to the sky in the first example of Fig.~\ref{fig:qualitative_4} and in the second example of Fig.~\ref{fig:qualitative_5}. {Some of these issues are fixed when using \methodShort\ to generate the vocabulary, as shown by the better quantitative results achieved in Tab. 4-6 by SAN + \methodShort.} Finally, compared to the SAM segmentation masks, the outputs presents slightly more noise on the edges, where regions are not perfectly segmented but are more approximate. 

Compared to the previous approaches, our method shows more coarse segmentation masks mainly due to the lack of any segmentation model to generate them. This is noticeable in, \eg, the first example of Fig.~\ref{fig:qualitative_4}, where the chimney and the sky have no clean separation, or in the second example of Fig.~\ref{fig:qualitative_5}, where the sky is not completely segmented as a single object. Moreover, it tends to propose larger amount of regions, \eg, first example of Fig.~\ref{fig:qualitative_5}, where the cat is separated into "cat", "ear", and "whisker", or in the second example of Fig.~\ref{fig:qualitative_6}, where the woman is segmented as "hair" and "shirt". Note, however, that \methodSShort\ is the only method to correctly classify and segment subtle details like the "cup"s in the left part of the second example of Fig.~\ref{fig:qualitative_6}, or the horse and grass in the first example of the same figure. Despite the coarse segmentation masks, our model provides more contexualised labels, \eg, "plane", "ceiling", "airline", "aisle", and "aircraft" for the airplane aisle in the second example of Fig.~\ref{fig:qualitative_4} (SAM and SAN selects labels such as "chair", "air travel", and "hair"), or "cat", "whisker", and "ear", in the first example of Fig.~\ref{fig:qualitative_5} (SAN selects "tabby", "gitana", "Freya", and "soporific" as labels). 

To conclude, all methods show their peculiarities, with SAM presenting the cleanest segmentation masks but sometimes undersegmenting objects, SAN having the most efficient implementation but showing sub-optimal performance with vocabularies out-of-distribution (\ie, outside of the common words used for semantic segmentation, as also shown by its weakness against "distractor" words), and \methodSShort\ recognising also smaller objects but resulting in the most coarse segmentation masks. {These results confirm the challenges of \taskSShort, and the need of developing methods tailored to this task.}

\vfill

\end{document}